\documentclass[nohyperref]{article}

\usepackage{natbib}
\usepackage{microtype}
\usepackage{graphicx}
\usepackage{subfigure}
\usepackage{adjustbox}
\usepackage{booktabs} % for professional tables

\usepackage{hyperref}

\usepackage[final]{neurips_2022}

\usepackage{bbm}

\newcommand{\R}{\mathbb{R}}
\newcommand{\E}{\mathbb{E}}
\newcommand{\fun}{\mathbf{f}}
\newcommand{\D}{\mathbf{D}}
\def\I{\mathbf{I}}
\newcommand{\act}{\mathbf{}{a}}
\usepackage{amsmath}
\usepackage{amssymb}
\usepackage{mathtools}
\usepackage{amsthm}
\usepackage[capitalize,noabbrev]{cleveref}

\theoremstyle{plain}
\newtheorem{theorem}{Theorem}[section]

\newtheorem{example}[theorem]{Example}

\theoremstyle{definition}

\newtheorem{hypothesis}[theorem]{Hypothesis}

\theoremstyle{remark}
\newtheorem{remark}[theorem]{Remark}

\usepackage[textsize=tiny]{todonotes}

\usepackage{neurips_2022}

\usepackage[utf8]{inputenc} % allow utf-8 input
\usepackage[T1]{fontenc}    % use 8-bit T1 fonts
\usepackage{hyperref}       % hyperlinks
\usepackage{url}            % simple URL typesetting
\usepackage{booktabs}       % professional-quality tables
\usepackage{amsfonts}       % blackboard math symbols
\usepackage{nicefrac}       % compact symbols for 1/2, etc.
\usepackage{microtype}      % microtypography
\usepackage{xcolor}         % colors

\title{First is Better Than Last for Language Data Influence}

\author{%
    Chih-Kuan Yeh \thanks{part of work done in CMU}  \\
  Google Inc.\\
  chihkuanyeh@google.com\\
    \And
    Ankur Taly  \\
  Google Inc.\\
  ataly@google.com\\
      \And
    Mukund Sundararajan\\
  Google Inc.\\
  mukunds@google.com\\
    \And
    Frederick Liu \\
  Google Inc.\\
  frederickliu@google.com\\
    \And
    Pradeep Ravikumar \\
  Carnegie Mellon University \\ Department of Machine Learning\\
  pradeepr@cs.cmu.edu\\
}

\begin{document}

\maketitle

\begin{abstract}
\vspace{-5mm}
    The ability to identify influential training examples enables us to debug training data and explain model behavior. Existing techniques to do so are based on the flow of training data influence through the model parameters~\citep{koh2017understanding, yeh2018representer, pruthi2020estimating}. For large models in NLP applications, it is often computationally infeasible to study this flow through \emph{all} model parameters, therefore techniques usually pick the last layer of weights. 
    However, we observe that since the activation connected to the last layer of weights contains ``shared logic'', the data influenced calculated via the last layer weights prone to a ``cancellation effect'', where the data influence of different examples have large magnitude that contradicts each other. The cancellation effect lowers the discriminative power of the influence score, and deleting influential examples according to this measure often does not change the model's behavior by much.
    To mitigate this, we propose a technique called TracIn-WE that modifies a method called TracIn~\citep{pruthi2020estimating} to operate on the word embedding layer instead of the last layer, where the cancellation effect is less severe. One potential concern is that influence based on the word embedding layer may not encode sufficient high level information. 
    However, we find that gradients (unlike embeddings) do not suffer from this, possibly because they chain through higher layers. We show that TracIn-WE significantly outperforms other data influence methods applied on the last layer significantly on the case deletion evaluation on three language classification tasks for different models. In addition, TracIn-WE can produce scores not just at the level of the overall training input, but also at the level of words within the training input, a further aid in debugging. 
\end{abstract}

\vspace{-7mm}
\section{Introduction}\label{sec:intro}
\vspace{-2mm}
Training data influence methods study the influence of training examples on a model's weights (learned during the training process), and in turn on the predictions of other test examples.
They enable us to debug predictions by attributing them to the training examples that most influence them, debug training data by identifying mislabeled examples, and fixing mispredictions via training data curation.
While the idea of training data influence originally stems from the study of linear regression ~\citep{cook1982residuals}, it has recently been developed for complex machine learning models like deep networks.

Prominent methods for quantifying training data influence for deep networks include influence functions~\citep{koh2017understanding}, representer point selection~\citep{yeh2018representer}, and TracIn~\citep{pruthi2020estimating}.
While the details differ, all methods involves computing the gradients (w.r.t. the loss) of the model parameters at the training and test examples.
Thus, they all face a common computational challenge of dealing with the large number of parameters in modern deep networks. In practice, this challenge is circumvented by restricting the study of influence to only 
%take gradient w.r.t. 
the parameters in the last layer of the network. While this choice may not be explicitly stated, it is often implicit in the implementations of larger neural networks. In this work, we revisit the choice of restricting influence computation to the last layer in the context of large-scale Natural Language Processing (NLP) models.

We first introduce the phenomenon of ``cancellation effect'' of training data influence, which happens when the sum of the influence magnitude among different training examples is much larger than the influence sum. This effect increases the influence magnitude of most training examples and reduces the discriminative power of data influence. We also observe that different weight parameters may have different level of cancellation effects, and the weight parameters of bias parameters and latter layers may have larger cancellation effects. To mitigate the ``cancellation effect'' and find a scalable algorithm, we propose to operate data influence on weight parameters with the least cancellation effect -- the first layer of weight parameter, which is also known as the word embedding layer.
%The representation generated after the word embedding layer is typically the first layer in any NLP model, which has not undergone any task-specific reductions or transformations. Thus, by design, it does not suffer from the shortcomings of last layer representations.

While word embedding representations might have the issue of not capturing any high-level input semantics, we surprisingly find that the gradients of the embedding weights do not suffer from this. Since the gradient chain through the higher layers, it thus takes the high-level information captured in those layers into account. As a result, the gradients of the embedding weights of a word depend on both the context and importance of the word in the input.
We develop the idea of word embedding based influence in the context of TracIn due to its computational and resource efficiency over other methods.
Our proposed method, TracIn-WE, can be expressed as the sum of word embedding gradient similarity over overlapping words between the training and test examples.
Requiring overlapping words between the training and test sentences helps capture low-level similarity%The sum over overlapping words captures low-level similarity between sentences
, while the word gradient similarity helps capture the high-level semantic similarity between the sentences.
A key benefit of TracIn-WE is that it affords a natural word-level decomposition, which is not readily offered by existing methods.
This helps us understand which words in the training example drive its influence on the test example.

We evaluate TracIn-WE on several NLP classification tasks, including toxicity, AGnews, and MNLI language inference with transformer models fine-tuned on the task.
We show that TracIn-WE outperforms existing influence methods on the case deletion evaluation metric by $4-10 \times$. A potential criticism of TracIn-WE is its reliance on word overlap between the training and test examples, which would prevent it from estimating influence between examples that relate semantically but not syntactically. To address this, we show that the presence of common tokens in the input, such as a ``start'' and ``end'' token (which are commonly found in modern NLP models), allows TracIn-WE to capture influence between semantically related examples without any overlapping words, and outperform last layer based influence methods on a restricted set of training examples that barely overlaps with the test example.\footnote{code is in \text{https://github.com/chihkuanyeh/TracIn-WE.}}

\vspace{-4mm}
\section{Preliminaries}\label{sec:preliminaries}
\vspace{-3mm}
Consider the standard supervised learning setting, with inputs $x \in \mathcal{X}$, outputs $y \in \mathcal{Y}$, and training data $\D = \{(x_1, y_1), (x_2, y_2), ... (x_n, y_n)\}$. Suppose we train a predictor $\fun$ with parameter $\Theta$ by minimizing some given loss function $\ell$ over the training data, so that $\Theta = \arg\min_{\Theta} \sum_{i=1}^n\ell(\fun(x_i), y_i)$. 
In the context of the trained model $\fun$, and the training data $\D$, we are interested in the data importance of a training point $x$ to the testing point $x'$, which we generally denote as $\I(x, x')$. 
\vspace{-3mm}
\subsection{Existing Methods}
\vspace{-3mm}
We first briefly introduce the commonly used training data influence methods: Influence functions~\citep{koh2017understanding}, Representer Point selection~\citep{yeh2018representer}, and TracIn~\citep{pruthi2020estimating}. 
%For each method, we demonstrate how , in large part due to the chain rule, all the data importance methods above
We demonstrate that each method 
can be decomposed into a similarity term $S(x,x')$, which measures the similarity between a training point $x$ and the test point $x'$, and loss saliency terms $L(x)$ and $L(x')$, that measures the saliency of the model outputs to the model loss.
The decomposition largely derives from an application of chain rule to the parameter gradients.
%As will be clear from our presentation below, in large part due to the chain rule, all the data importance methods above can be decomposed into a similarity term, which measures the similarity between a training point $x$ and the test point $x'$, and a loss saliency term, that measures the saliency of the model outputs to the model loss.
\[\I(x, x') = L(x) S(x,x') L(x')\]
% where the label information is only related to the loss term (expect $H_\Theta$ in influence function).
%In each case, the decomposition can be derived by applying the chain rule to parameter gradients.
%The intuition of the decomposition is natural: 
The decomposition yields the following interpretation.
A training data $x$ has a larger influence on a test point $x'$ if (a) the training point model outputs have high loss saliency, (b) the training point $x$ and the test point $x'$ are similar as construed by the model.
In Section~\ref{sec:last},we show that restricting the influence method to operate on the weights in the last layer of the model critically affects the similarity term, and in turn the quality of influence.
We now introduce the form of each method, and the corresponding similarity and loss saliency terms. 
%$L(x)$, $L(x')$,  and the similarity terms $S(x,x')$.

%, as well as the formulation when the method is adapted to the final parameter layer of the model.
\vspace{-2mm}
\paragraph{Influence Functions:} 
\vspace{-2mm}
\[{\text{Inf}}(x, x') = -\nabla_{\Theta} \ell(x, \Theta)^T H_{\Theta}^{-1} \nabla_{\Theta} \ell(x', \Theta),\]
where $ H_{\Theta}$ is the hessian  $\sum_{i=1}^n \nabla^2_{\Theta} \ell(x, \Theta)$ computed over the training examples. By an application of the chain rule, we can see that $\text{Inf}(x, x') = L(x) S(x,x') L(x')$, with the similarity term $S(x,x') = \frac{\partial\fun(x, \Theta)}{\partial \Theta}^T H_{\Theta}^{-1}\frac{\partial\fun(x', \Theta)}{\partial \Theta}$, and the loss saliency terms $L(x) =  \frac{\partial \ell(x, \Theta)}{\partial \fun(x, \Theta)}$. The work by \citet{sui2021representer} is very similar to extending the influence function to the last layer to satisfy the representer theorem.

%Since the construction involves taking gradients with respect to model parameters, one could compute these only with respect to some, rather than all, parameters. When applied to the parameter of the last layer, we get:
%\[\I_{\text{Inf-Last}}(x, x') = -\nabla_{\Theta_\text{last}} \ell(x, \Theta)^T H_{\Theta_{\text{last}}}^{-1} \nabla_{\Theta_{\text{last}}} \ell(x', \Theta),\]
\vspace{-1mm}
\paragraph{Representer Points:} 
\vspace{-2mm}
\begingroup
\setlength\abovedisplayskip{-2mm}
\setlength\abovedisplayskip{-0.5mm}
\begin{equation} 
\text{Rep}(x, x') = -\frac{1}{2\lambda n} \frac{\partial \ell(x, \Theta)}{\partial \fun_j(x, \Theta)}\act(x, \Theta)^T \act(x', \Theta),
\end{equation}
\endgroup
where $\act(x, \Theta)$ is the final activation layer for the data point $x$, $\lambda$ is the strength of the $\ell_2$ regularizer used to optimize $\Theta$, and $j$ is the targeted class to explain. The similarity term is $S(x,x') = 
\act(x, \Theta)^T \act(x', \Theta)$,
% \frac{\partial\fun_j(x, \Theta)}{\partial \Theta}^T \frac{\partial\fun_j(x', \Theta)}{\partial \Theta}$, 
and the loss saliency terms are $L(x) = \frac{1}{2\lambda n} \frac{\partial \ell(x, \Theta)}{\partial \fun_j(x, \Theta)}$, $L(x') = 1.$ %The method is already implicitly targeting the parameter of the final layer, since the final layer activations can be written as $\act(x, \Theta)= \frac{\partial \fun(x,\Theta)}{\partial \Theta_{\text{last}}}$.
\vspace{-2mm}
\paragraph{TracIn:} 
\vspace{-2mm}
\begingroup
\setlength\abovedisplayskip{-3mm}
\setlength\abovedisplayskip{-1mm}
\begin{equation} 
\text{TracIn}(x, x') = - \sum_{c = 1}^d  \eta_c  \nabla_{\Theta_c} \ell(x, \Theta_c)^T \nabla_{\Theta_c} \ell(x', \Theta_c),
\end{equation}
\endgroup
%where
%\[\I_{\text{Tr}}(x, x', c) = - \nabla_{\Theta_c} \ell(x, \Theta_c)^T \nabla_{\Theta_c} \ell(x', \Theta_c),\]
where $\Theta_c$ is the weight at checkpoint $c$, and $\eta_c$ is the learning rate at checkpoint $c$. In the remainder of the work, in our notation, we suppress the sum over checkpoints of TracIn for notational simplicity.
(This is not to undermine the importance of summing over past checkpoints, which is a crucial component in the working on TracIn.)
For TracIn, the similarity term is $S(x,x') =  \nabla_{\Theta} \fun(x, \Theta)^T \nabla_{\Theta} \fun(x', \Theta)$, while the loss terms are $L(x) =  \frac{\partial \ell(x, \Theta)}{\partial \fun(x, \Theta)}$, $L(x') =  \frac{\partial \ell(x', \Theta)}{\partial \fun(x', \Theta)}$. 
\vspace{-4mm}
\subsection{Evaluation: Case Deletion}\label{sec:eval_metric}
\vspace{-3mm}
%It is instructive at this juncture to also
We now discuss our primary evaluation metric, called \emph{case deletion diagnostics}~\citep{cook1982residuals}, which involves retraining the model after removing influential training examples and measuring the impact on the model. This evaluation metric helps validate the efficacy of any data influence method in detecting training examples to remove or modify for targeted fixing of misclassifications, which is the primary application we consider in this work. This evaluation metric was also noted as a key motivation for influence functions~\citep{koh2017understanding}.
Given a test example $x'$, when we remove training examples with positive influence on $x'$ (\emph{proponents}), we expect the prediction value for the ground-truth class of $x'$ to decrease. On the other hand, when we remove training examples with negative influence on $x'$ (\emph{opponents}), we expect the prediction value for the ground-truth class of $x'$ to increase. 

An alternative evaluation metric is based on detecting mislabeled examples via self-influence (i.e. influence of a training sample on that same sample as a test point). We prefer the case deletion evaluation metric, as it more directly corresponds to the concept of data influence. Similar evaluations that measure the change of predictions of the model after a group of points is removed is seen in previous works.~\citet{han2020explaining} measures the test point prediction change after $10\%$ training data with the most and least influence are removed, and ~\citet{koh2019accuracy} measures the correlation of the model loss change after a group of trained data is removed and the sum of influences of samples in the group, where the group can be seen as manually defined clusters of data.
%We contrast the case deletion evaluation to a commonly-used evaluation: detecting mislabeled example by self-influence. The main difference is that the deletion evaluation directly measures how removing a training point will affect a test point, while the self-influence evaluation relies on the implicit assumption that mislabeled examples would have large self-influence. While the computational cost for the case deletion evaluation is high compared to the self-influence evaluation, the case deletion is a more and should be preferred when the computational aspect is affordable.
%The traditional data influence originated from the study of linear regression ~\citep{cook1982residuals}, which suggested that case deletion diagnostics have found the greatest acceptance at the time. Case deletion refers to removing the training data and retrain the model, and is also used as the motivating argument for \cite{koh2017understanding}. When we remove a positively related training data $x$ (proponents), we expect the prediction value of the test point $x'$ to decrease. On the other hand, when we remove a negatively related training data $x$ (opponents), we expect the prediction value of the test point $x'$ to increase. Therefore, given a testing point $x'$, we define the metric $\textsc{del}_{+}(x')$ and $\textsc{del}_{-}(x')$ as the following:

\vspace{-3mm}
\paragraph{Deletion curve.}
Given a test example $x'$ and influence measure $\I$, we define the metrics $\textsc{del}_{+}(x', k,\I)$ and $\textsc{del}_{-}(x',k,\I)$ as the impact on the prediction of $x'$ (for its groundtruth class) upon removing top-$k$ proponents and opponents of $x'$ respectively:
\[\textsc{del}_{+}(x',k, \I) = \E[f_c(x', \Theta_{+;k}) - f_c(x', \Theta) ], \]
\[\textsc{del}_{-}(x',k, \I) = \E[f_c(x', \Theta_{-;k}) - f_c(x', \Theta) ], \]
where, $\Theta_{+;k}$ ( $\Theta_{-;k}$) are the model weights learned when top-$k$ proponents (opponents) according to influence measure $I$ are removed from the training set, and $c$ is the groundtruth class of $x'$. 
The expectation is over the number of retraining runs.
%We use $10-$run average to estimate the expected value in the context of this work.
%\footnote{We remark that the metrics $\textsc{del}_{+}(x')$ and $\textsc{del}_{-}(x')$ can also be defined by the difference of losses instead of model outputs. One practical issue with that choice however is that the average loss is often dominated by a few mis-predictions, which is not ideal when only a small set of test examples can be chosen since this evaluation is expensive.} 
We expect  $\textsc{del}_{+}$ to have large negative, and $\textsc{del}_{-}$ to have large positive values. To evaluate the deletion metric at different values of $k$, we may plot $\textsc{del}_{+}(x',k, \I)$ and $\textsc{del}_{-}(x',k, \I)$ for different values of $k$, and report the area under the curve (AUC): $\textsc{auc-del}_+ = \sum_{k=k_1}^{k_m} \frac{1}{m}\textsc{del}_{+}(x',k, \I)$, and $\textsc{auc-del}_- = \sum_{k=k_1}^{k_m} \frac{1}{m}\textsc{del}_{-}(x',k, \I)$.

We note that the \emph{case deletion diagnostics} is different to the leave-one-out evaluation of ~\citet{koh2017understanding} by two points. First, leave-one-out evaluation focuses on removing one point, which is more meaningful in the convex regime where the optimization is initialization-invariant. We consider the leave-k-out evaluation which is closer to actual applications, as one may need to alter more than one training data to fix a prediction. Second, we consider the expected value of leave-k-out, to hedge the variance caused by specific model states, which was pointed out by \citet{sogaard2021revisiting} to be a major issue for leave-one-out evaluation (especially when the objective is no longer convex).

\vspace{-4mm}
\section{Cancellation Effect of Data Influence} \label{sec:cancel}
\vspace{-3mm}
The goal of a data influence method is to distribute the test data loss (prediction) across training examples, which can be seen as an attribution problem where each training example is an agent.
We observe \emph{cancellation} across the data influence attributions to training examples,
i.e., the sign of attributions across training examples disagree and cancels each other out.
%This occurs when the magnitude of the attributions sum to much more than the actual influence.
%This happens when the sign of attributions across agents disagree and cancels each other out.
This leads to most training examples having a large attribution magnitude, which reduces the discriminatory power of  attribution-based explanations. 

Our next observation is that the cancellation effect varies across different weight parameters.
In particular, when a weight parameter is used by most of the training examples, the cancellation effect is especially severe.
One such parameter is the bias, whose cancellation effect is illustrated by the following example:

\begin{example} \label{ex:cancel}
Consider an example where the input $x \in \R^d$ is sparse, and $x_i$ has feature $i$ with value 1 and all other features with value 0. 
The prediction function has the form $\fun(x) = x \cdot w + b$.
It follows that a set of optimal parameters are $w_i = y_i, b=0$.
We further assume that the parameter $b$ is initialized to $0$ and has never changed during the gradient descent progress. In this case, it is clear that the bias parameter $b$ is irrelevant to the model (as removing it will not change the model at all). However, the influence to individual examples caused by the bias $b$ is still non-zero. This is because even that the sum of gradient for bias $b$ is $0$, (s.t. $\sum_i \frac{\partial L(\fun(x_i),y_i)}{\partial b} = 0$), each individual term $\frac{\partial L(\fun(x_i),y_i)}{\partial b} = \frac{\partial L(\fun(x_i),y_i)}{\partial \fun(x_i)}$ is non-zero for most $x_i$. Note that $\frac{\partial L(\fun(x_i),y_i)}{\partial b}$ contributes to the influence to data $x_i$ directly, and thus the bias parameter $b$ will contribute to the influence of all the training data. In the contrary, for each weight variable $w_j$, $\frac{\partial L(\fun(x_i),y_i)}{\partial w_j}$ is only non-zero for $x_j$, and thus the weight variable $w_j$ only contributes to the influence of one training data $x_j$. Thus, the bias would affect the influence for more training examples compared to the weights.
\end{example}
\vspace{-2mm}

The above example illustrates that while the bias parameter is not important for the prediction model (removing the bias can still lead to the same optimal solution), the total gradient that flows through the bias still high. In fact, we find empirically that the total influence that flows through the bias is larger than that flowing through the weight, since each training example's gradient will affect the bias but the total contribution will be cancelled out, so the bias will remain $0$.
We also note that even for deep network models that do not have a sparse input, the neurons connected to the weight are often $0$ (due to ReLU types of activation functions). Thus, the gradient of weight parameters is often sparser compared to the gradient of bias parameters, and thus bias parameters would often have stronger cancellations, which we validate empirically.
\vspace{-4mm}
\subsection{Measuring the Cancellation Effect}
\vspace{-3mm}
In the above example, we defined strong cancellation effect when some weight parameters does not change a lot during training (or has saturated in the training process), but the total strength of the gradient of the weight parameters summed over training data is large. For weights $W$, we first define two terms $\Delta W_c$ and $G(W)_c$, 
$$\Delta W_c = \|W_{c+1} - W_c\|,$$
$$ G(W)_c = \sum_{x_i,y_i \sim D} \eta_c \|\frac{\partial l(x_i,y_i)}{ \partial {W_c}}\|,$$

where $\Delta W^c$ measures the norm of weight parameter change between checkpoint $c$ and $c+1$, and $G(W)_c$ measures the sum of weight gradient norm times learning rate summed over all training data. When $\Delta W_c$ is small, this means that the weight $W$ may have saturated at checkpoint $c$, and the weight may not actually affect the model output much (and thus the weight $W$ is not important for this epoch of training). When $G(W)_c$ is large, this means that the sum of gradient norm with respect to $W_c$ is still large, and the influence norm caused by $\frac{\partial l(x_i,y_i)}{ \partial {W_c}}$ will also be large.

To measure the cancellation effect, we define the cancellation ratio of a weight parameter $W$ as:
\vspace{-1mm}
\[C(W) = \frac{\sum_{c}G(W)_c}{\sum_{c}\Delta W_c}.\]
When $G(W)_c$ is large and $\Delta W_c$ is small, this means that a non-important weight $W_c$ greatly influenced the total influence norm, which may be only possible if the influence contributed from $W_c$ to different examples cancelled each other out. Applying this interpretation to the cancellation of bias parameters, the intuition is that the bias parameters are not mainly responsible for the reduction of testing example loss change during the training process (since $\Delta W_c$ is small). However, they dominate the total influence strength due to their dense nature ($G(W)_c$  is large). Parameters with high cancellation may not be ideal to the calculation of influences.

\vspace{-4mm}
\subsection{Removing Bias In TracIn Calculation to Reduce Cancellation Effect}
\vspace{-3mm}
To investigate whether removing weights with high cancellation effect  really helps improve influence quality,  we conducted an experiment on a CNN text classification on Agnews dataset with $87 \%$ test accuracy. 
%To investigate if our  first layer is \emph{only} working since it's only been applied to fine-tuned models, we have conducted an experiment on a CNN text classification on Agnews dataset with $87 \%$ test accuracy. 
The model is defined as follows: first a token embedding with dimension $128$, followed by two convolution layers with kernel size $5$ and filter size $10$, one convolution layers with kernel size $1$ and filter size $10$, a global max pooling layer, and a fully connected layer; all weights are randomly initialized. The first layer is the token embedding, the second layer is the convolution layer, and the last layer is a fully connected layer. 
The model has $21222$ parameters in total (excluding the token embedding), in which $102$ parameters are bias variables. 
%The average total absolute gradient passing through each variable $\sum_{x_i \in \mathcal{X}} \sum _{w \in \mathcal{W}} \frac{1}{21222}|\frac{\partial L(\fun(x_i),y_i)}{\partial w}|$ is $660.44$. On the other hand, the average total absolute gradient passing through each bias variable $b$, which can be written as $\sum_{x_i \in \mathcal{X}} \sum _{b \in \mathcal{B}} \frac{1}{102}|\frac{\partial L(\fun(x_i),y_i)}{\partial b}|$ is $13396.24$. Clearly, much more gradient flows through the bias parameters compared to non-bias parameters. 
%We decompose the TracIn score to TracIn-bias and TracIn-weight
%\vspace{-2mm}
%\[\text{TracIn-bias}(x, x') = - \sum_{c = 1}^d  \eta_c  \nabla_{\Theta_{b,c}} \ell(x, \Theta_c)^T \nabla_{\Theta_{b,c}} \ell(x', \Theta_c),\]
%\vspace{-3mm}
%\[\text{TracIn-weight}(x, x') = - \sum_{c = 1}^d  \eta_c  \nabla_{\Theta_{w,c}} \ell(x, \Theta_c)^T \nabla_{\Theta_{w,c}} \ell(x', \Theta_c),\]
%which are the TracIn scores contributed from the bias parameters $\Theta_b$ and weight parameters $\Theta_{w}$ respectively. 
We find  $C($bias$)$ to be $16789$, and $C($weight$)$ to be $2555$, which validates that the bias variables have a much stronger cancellation effect than the weight variables. A closer analysis shows that $G(\text{bias})$ is similar to $G(\text{weight})$ ($627206$ and $559142$), but $\Delta(\text{bias})$ is much smaller than $\Delta(\text{weight})$ ($0.74$ and $4.37$.) Even though the bias parameters has a much smaller total change compared to the weight parameters, their impact on the gradient norm (and thus influence norm) is even higher than the weight parameters. This verifies the intuition in Example \ref{ex:cancel} that the bias parameter has a stronger cancellation effect since the gradient to bias is almost activated for all examples despite the actual bias change being small.
To further verify that the TracIn score contributed by the bias may lower the overall discriminatory power, we compute $\textsc{auc-del}_+$ and $\textsc{auc-del}_-$ for TracIn and TracIn-weight on AGnews with our CNN model. The $\textsc{auc-del}_+$ for TracIn and TracIn-weight is $-0.036$ and $-0.065$ respectively, and the $\textsc{auc-del}_+$ for TracIn and TracIn-weight is $0.011$ and $0.046$. The result shows that by removing the TracIn score contributed by the bias (with only $102$ parameters), the overall influence quality improves significantly. Thus, in all future experiments, we remove the bias in calculation of data influence if not stated otherwise.

\iffalse
We compute TracIn-WE, TracIn-Sec (applied on second layer), TracIn-Last, and TracIn-All (applied on all layers) for number of removed examples $k \in [30,70]$, and average over $20$ test points. The AUC-DEL+ (lower is better) for TracIn-WE, TracIn-Sec, TracIn-Last, TracIn-All respectively is $-0.077, -0.075, -0.016, -0.065$, and AUC-DEL- (higher is better) is $0.045, 0.022, 0.014, 0.047$. Observe that: (1) TracIn-WE works much better than TracIn-Last even when trained from scratch, (2) TracIn-WE is on-par/ slightly better than TracIn-All, showing that the first layer alone may be sufficient.
\vspace{-2mm}
\paragraph{Is representation layer crucial? (R6) Does it have to be a word embedding layer? (R1):} 
As reported above, for CNN text classification, both the first layer (token embedding mentioned by R6) and the second layer work well compared to the last layer. In Appendix F, we report that the second layer works well for a Bert model as well.
Observe that the second layer in Bert and in CNN, are neither representation layers nor word embedding layers, but they work reasonably well and better than the last layer.
We believe the critical reason is that the gradient of earlier layers do not suffer from the information collapse issue compared to the last layer.
\fi

\vspace{-4mm}
\subsection{Influence of Latter Layers May Suffer from Cancellation} \label{sec:last}
\vspace{-3mm}

As mentioned in Section
~\ref{sec:intro}, for scalability reasons, 
most influence methods choose to operate only on the parameters of the last fully-connected layer $\Theta_\text{last}$.
We argue that this is not a great choice, as the influence scores that stems from the last fully-connected weight layer may suffer from cancellation effect, as different examples ``share logics'' in the activation representation of this layer, and have a higher gradient similarity for different examples. 
Early layers, where examples have unique logic, may suffer less from the cancellation effect. 
We first measure the gradient similarity for different examples for each layer, which is 
$E_{x_a,x_b} {\text{COS-SIM} [\partial l(x_a)/ \partial w , \partial l(x_b)/ \partial w]}$, where $\text{COS-SIM}$ is the cosine similarity. This measures the expected gradient cosine similarity between two examples. The expected gradient similarity for testing examples between different layers in the CNN classification are: first $0.035$, second $0.075$, third $0.21$, last $0.23$. This verifies that the latter layers in the neural network have more aligned gradients between examples, and thus share more logics between training examples.
We report the cancellation ratio for each of the TracIn layer varaint in Table \ref{tab:auc-agnews}, where TracIn-first, TracIn-second, TracIn-third, TracIn-last, TracIn-All refer to TracIn scores based on weights of the first layer, second layer, third layer, last layer, and all layers (the bias is always omitted). As we suspected, early layers suffers less from cancellation, and latter layers suffers more from cancellation.
To assess the impact on influence quality, we evaluate the $\textsc{auc-del}_+$ and $\textsc{auc-del}_-$ score for TracIn calculated with different layers on the AGnews CNN model in Tab.~\ref{tab:auc-agnews}.
We observe that removing examples based on influence scores calculated using parameters of later layers (with more ``shared logic"") leads to worse deletion score compared to removing examples based on influence scores calculated using parameters of earlier layers (with more ``unique logic'').
Interestingly, the performance of TracIn-first even outperforms TracIn-all where all parameters are used. \footnote{We note that our investigation of last layer cancellation is limited to the setting when the whole model is trained to produce a single classification score, which may not hold in the setting where only the last layer is fine-tuned or tasks with a generative output.}
We hypothesize that since the TracIn score based on later layers contain too much cancellation, it is actually harmful to include these weight parameters in the TracIn calculation. 
In the following, we develop data influence methods by only using the first layer of the model, which suffers the least from cancellation effect.

 \begin{table*}[t!]
    \centering% This is an environment - we probably don't want the extra spacing of center in addition to that added by table etc.
    \caption{Cancellation Ratio and AUC-DEL table for various layers in CNN model in AGnews. }
    \vspace{1mm}
    \label{tab:auc-agnews}
    \small
    \begin{tabular}{p{0.08\textwidth} p{0.15\textwidth} p{0.10\textwidth}p{0.10\textwidth}p{0.10\textwidth}p{0.10\textwidth}p{0.10\textwidth}}
      \toprule% nicer rules courtesy of booktabs - but then we need to drop the verticals 
      Dataset & Metric & TR-first & TR-second & TR-third& TR-last& TR-all \\
      \midrule
      AGnews & Cancellation$\ \downarrow$ & $\ \ \mathbf{1863}$ & $\ \ {2019}$ & $\ \ 3126$& $\ \ 2966$ & $\ \ 2368$ \\
      & AUC-DEL$+\downarrow$ & $\mathbf{-0.077}$ & $\mathbf{-0.075}$ & $0.012$& $-0.016$ & ${-0.065}$ \\
       & AUC-DEL$-\uparrow$ & $\ \ \ \mathbf{0.045}$ & $\ \ \ 0.022$ & $ 0.006$& $ {-0.032}$ & $\ \ \ \mathbf{0.046}$ \\
      \midrule
      \bottomrule
      %\vspace{-5mm}
    \end{tabular}
    \vspace{-2.5mm}
  \end{table*}

\vspace{-4mm}
\section{Word Embedding Based Influence} \label{sec:TracIn_we}
\vspace{-3mm}
In the previous section, we argue that using the latter layers to calculate influence may lead to the cancellation effect, which over-estimates influence.
Another option is to calculate influence on all weight parameters, but may be computational infeasible when larger models with several millions of parameters are used.
% Consequently, influence methods operating on the last layer result in influential examples that are  unintuitive and do not do well on the case deletion evaluation.
%The main shortcoming of last layer representations is that they are too high-level, which makes similarity in this space collapse to label similarity.
To remedy this, we propose operating on the first layer of the model, which contains the less cancellation effect since early layers encodes ``unique logit''. The first layer for language classification models is usually the word embedding layer in the case of NLP models. However, there are two questions in using the first layer to calculate data influence: 1. the word (token) embedding contains most of the weight parameters, and may be computational expensive 2. the word embedding layer may not capture influential examples through high-level information.
%Surprisingly, we find that gradients to the word embedding layer can capture both high-level and low-level information about the input sentence, and therefore serve as a appropriate representation for computing similarity.
In the rest of this section, we develop the idea of word embedding layer based training-data influence in the context of TracIn. We focus on TracIn due to challenges in applying the other methods to the word embedding layer: influence functions on the word embedding layer are computationally infeasible due to the large size (vocab size $\times$ embedding\_dimension) of the embedding layer, and representer is designed to only use the final layer. We show that our proposed influence score is scalable thanks to the sparse nature of word embedding gradients, and contains both low-level and high-level information since the gradient to the word embedding layer can capture both high-level and low-level information about the input sentence.

\vspace{-2mm}
 \begin{table*}[t]
    \centering% This is an environment - we probably don't want the extra spacing of center in addition to that added by table etc.
    \caption{Examples for word similarity for different examples containing word ``not''. }
    \label{tab:word_sim}
    \small
    \adjustbox{max width=0.9\columnwidth}{
    \begin{tabular}{p{0.07\textwidth} p{0.40\textwidth}  p{0.35\textwidth} p{0.1\textwidth}}
      \toprule% nicer rules courtesy of booktabs - but then we need to drop the verticals 
      Example & Premise & Hypothesis & Label \\\midrule% Note that there is no & before the first column - & only comes between columns so if you define n columns, you can have at most n-1 & symbols in any row
      S1 & I think he is very annoying. & I do \textbf{not} like him.& Entailment \\% No need for 6 rows of space for each title!
      S2 & I think reading is very boring. & I do \textbf{not} like to read.& Entailment \\% No need for 6 rows of space for each title!
      S3 & I think reading is very boring. & I do \textbf{not} hate burying myself in books.& Contradiction \\% No need for 6 rows of space for each
      S4 & She \textbf{not} only started playing the piano before she could speak, but her dad taught her to compose music at the same time. & She started to playing music and making music from very long ago.&  Entailment\\% No need for 6 rows of space for each title!
      \midrule 
      S5 & I think he is very annoying. & I don't like him.& Entailment \\
      S6 & She thinks reading is pretty boring & She doesn't love to read& Entailment \\
      S7 & She not only started playing the piano before she could speak, but her dad taught her to compose music at the same time & She started to playing music and making music from quite long ago & Entailment \\

      \bottomrule
    \end{tabular}}
     \vspace{-5mm}
  \end{table*}

 %We have shown that when applying the influence methods on the last layer of a neural network, the resulting influence method suffers from the collapse of embedding similarity. When we replace the embedding similarity by common sentence similarity such as TF-IDF, the deletion curve results greatly improved. However, TF-IDF similarity does not take the model parameters into account. Since applying the influence methods on the later layer leads to information collapse, it is natural to consider applying these data influence methods on the first layer of the model.

%Since this work focuses on NLP applications, the first layer of the NLP model is usually the word embedding layer which transforms the input of word tokens into sentence of word embedding. Also, the word embedding layer is universally existent in various different deep language model architecture, it would be a safe layer to choose in practice. Since the word embedding is one of the largest layers for general deep learning models for NLP, computing the influence function of the word embedding is pretty computational infeasible, and representer is designed to only use the final layer. Thus, we propose to apply TracIn on the word embedding layer, which can be implemented efficiently due to the sparsity nature of the word embedding gradient.
\vspace{-2mm}
\subsection{TracIn on Word Embedding Layer}\label{sec:TracInwe}
\vspace{-3mm}
We now apply TracIn on the word embedding weights, obtaining the following expression:
\vspace{-2mm}
\begin{equation} \label{eq:trackemb}
\begin{split}\small
    \text{TracIn-WE}(x,x') = - \frac{\partial \ell(x, \Theta)}{\partial \Theta_{\text{WE}}} ^T \frac{\partial \ell(x', \Theta)}{\partial \Theta_{\text{WE}}},
\end{split}
\end{equation}
Implementing the above form of TracIn-WE would be computationally infeasible as word embedding layers are typically very large (vocab size $\times$ embedding dimension). For instance, a BERT-base model has $23$M parameters in the word embedding layer. 
To circumvent this, we  leverage the sparsity of word embedding gradients $\frac{\partial \ell(x, \Theta)}{\partial \Theta_{\text{WE}}}$, which is a sparse vector where only embedding weights associated with words that occur in $x$ have non-zero value. Thus, the dot product between two word embedding gradients has non-zero values only for words $w$ that occur in both $x, x'$. With this observation, we can rewrite TracIn-WE as:
\vspace{-4mm}

\begingroup
\setlength\abovedisplayskip{-2mm}
\setlength\belowdisplayskip{-3mm}
   \begin{equation} \label{eq:trackemb_we}
\begin{split}\small
    \text{TracIn-WE}(x,x') = - \sum_{w \in  x \cap x'} \frac{\partial \ell(x)}{\partial \Theta_{w}}^T \cdot \frac{\partial \ell(x')}{\partial \Theta_{w}},
\end{split}
\end{equation}
\endgroup

where $\Theta_{w}$ are the weights of the word embedding for word $w$. We call the term $\frac{\partial \ell(x)}{\partial \Theta_{w}}^T \cdot \frac{\partial \ell(x')}{\partial \Theta_{w}}$ the \emph{word gradient similarity} between sentences $x, x'$ over word $w$.
\vspace{-3mm}

 \begin{table*}[ht]
 \vspace{-2mm}
    \centering% This is an environment - we probably don't want the extra spacing of center in addition to that added by table etc.
    \caption{Word Decomposition Examples for TracIn-WE }
    \label{tab:decom}
    \small
    \adjustbox{max width=\columnwidth}{
    \begin{tabular}{p{0.19\textwidth} p{0.72\textwidth} p{0.1\textwidth}}
      \toprule
      & Sentence content & Label \\
      \toprule% nicer rules courtesy of booktabs - but then we need to drop the verticals 
      Test Sentence 1 - T1 & I can always end my conversations so you would not get any answers because you are too lazy to remember anything & Toxic  \\\midrule% Note that there is no & before the first column - & only comes between columns so if you define n columns, you can have at most n-1 & symbols in any row
      Test Sentence 2 - T2 & For me, the lazy days of summer is not over yet, and I advise you to please kindly consider to end one's life, thank you & Toxic\\\midrule% No need for 6 rows of space for each title!
      Train Sentence - S1 & Oh yeah, if you're too lazy to fix tags yourself, you're supporting AI universal takeover in 2020. end it. kill it now. & Non-Toxic\\\midrule % No need for 6 rows of space for each title!\\
      &Word Importance& Total \\ \midrule
      \small
      TracIn-WE(S1, T1) & [S]: $-0.28$, [E]: $-0.07$,  to: $-0.15$, \textbf{lazy}: $-7.6$, you: $-0.3$, end:$-0.3$, too:$-0.3$& $-9.2$ \\\midrule% No
      TracIn-WE(S1, T2) & [S]: $-0.17$, [E]: $-0.23$,   to: $0.54$, lazy: $-0.25$, you: $0.25$, \textbf{end}: $-3.12$ & $-3.45$ \\\midrule% No
      \bottomrule
    \end{tabular}}
    \vspace{-5mm}
  \end{table*}

\vspace{-3mm}
\subsection{Interpreting Word Gradient Similarity} \label{sec:wgs}
\vspace{-3mm}
Equation~\ref{eq:trackemb_we} gives the impression that TracIn-WE merely considers a bag-of-words style similarity between the two sentences, and does not take the semantics of the sentences into account.
This is surprisingly not true!
Notice that for overlapping words, TracIn-WE considers the similarity between gradients of word embeddings.
Since gradients are back-propagated through all the intermediate layers in the model, they take into account the semantics encoded in the various layers. 
This is aligned with the use of word gradient norm $\|\frac{\partial \fun(x)}{\partial \Theta_{w}}\|$ as a measure of importance of the word $w$ to the prediction $\fun(x)$~\citep{wallace2019allennlp, simonyan2013deep}.
Thus, word gradient similarity would be larger for words that are deemed important to the predictions of the training and test points.

Word gradient similarity is not solely driven by the importance of the word.
Surprisingly, we find that word gradient similarity is also larger for overlapping words that appear in similar contexts in the training and test sentences. We illustrate this via an example.
Table \ref{tab:word_sim} shows 4 synthetic premise-hypothesis pairs for the  Multi-Genre Natural Language Inference (MNLI) task~\citep{N18-1101}.
%We take an existing pretrained model to motivate our example [add: Deberta model from hugging face].
%Examples S1, S2, S4 has GT label entailment, while example S3 has GT label contradiction. 
An existing pretrained model~\citep{he2020deberta} predicts these examples correctly with softmax probability between $0.65$ and $0.93.$ 
Notice that all examples contain the word `not' once. The word gradient importance $\|\frac{\partial \fun(x)}{\partial \Theta_{w}}\|$ for ``not'' is comparable in
all $4$ sentences.
%$[0.5,1]$ for $4$ sentences, showing that the importance of the word ``not'' is similar in the examples.
The value of word gradient similarity for `not' is $0.34$ for the pair S1-S2, and $-0.12$ for S1-S3, while it is $-0.05$ for S1-S4.
This large difference stems from the context in which `not' appears. 
The absolute similarity value is larger for S1-S2 and S1-S3, since `not' appears in a negation context in these examples.
(The word gradient similarity of S1-S3 is negative since they have different labels.) %but same context.
However, in S4, `not' appears in the phrase ``not only ... but'', which is not a negation (or can be considered as double negation). Consequently, word gradient similarity for `not' is small
% (and slightly negative)
between S1 and S4. 
In summary, we expect the absolute value of TracIn-WE score to be large for training and test sentences that have overlapping important words in similar (or strongly opposite) contexts.
On the other hand, overlap of unimportant words like stop words would not affect the TracIn-WE score.
%We posit that the high-level information that is captured by word gradient similarity stems from the fact that the gradient flow of word embedding follows through all intermediate layers of the model.
\vspace{-4mm}
\subsection{Word-Level Decomposition for TracIn-WE}\label{sec:word_decomposition}
\vspace{-3mm}
An attractive property of TracIn-WE is that it decomposes into word-level contributions for both the testing point $x'$ and the training point $x$.
As shown in ~\eqref{eq:trackemb_we}, word $w$ in $x$ contributes to $\text{TracIn-WE}(x,x')$ by the amount $\frac{\partial \ell(x)}{\partial \Theta_{w}}^T \cdot \frac{\partial \ell(x')}{\partial \Theta_{w}} \mathbbm{1}[w \in x']$; a similar word-level decomposition can be obtained for $x'$. 
Such a decomposition helps us identify which words in the training point ($x$) drive its influence towards the test point ($x'$).
For instance, consider the example in Table.~\ref{tab:decom}, which contains two test sentences (T1, T2) and a training sentence S1. We decompose the score TracIn-WE(S1, T1) and TracIn-WE(S1,T2) into words contributions, and we see that the word ``lazy'' dominates TracIn-WE(S1, T1), and the word ``end'' dominates TracIn-WE(S1, T2). This example shows that different key words in a training sentence may drive influence towards different test points.
The feature-decomposition for influence introduces additional interpretability to why two examples are highly influenced. This is demonstrated in a case study where we cluster difficult training examples based on a normalized TracIn-WE score in Sec. \ref{sec:cluster}.
%We demonstrate such an application in Section \ref{sec:application}.
%we can replace the keywords in $x$ that contributes negatively to $x'$ by a similar word or simply remove the most influential key words in, as we will test in our experiments.

\vspace{-5mm}
\subsection{An approximation for TracIn-WE} \label{sec:computational_tricks}
\vspace{-3mm}
As we note in Sec. \ref{sec:TracInwe}, the space complexity of saving training and test point gradients scales with the number of words in the sentence.
This may be intractable for tasks with very long sentences. We alleviate this by leveraging the fact that the word embedding gradient for a word $w$ is the sum of input word gradients from each position where $w$ is present.
%In other words, 
%\[\frac{\partial\fun(x, \Theta)}{\partial \Theta_{w}} = \sum_{x^i = w}\frac{\partial\fun(x, {\Theta})}{\partial {x^i}}\]
%where $x^i$ is the word at position $i$ in input $x$.
Given this decomposition, we can approximate the word embedding gradients by saving only the top-k largest input word gradients for each sentence. (An alternative is to save the input word gradients that are above a certain threshold.)
%Formally, the approximation $\left\frac{\partial \ell(x, \Theta)}{\partial \Theta_{w}}\right\vert_{top-k}$ is defined as $\sum_{x^i = w}\frac{\partial \ell(x, {\Theta})}{\partial {x^i}} \mathbbm{1}[\|\frac{\partial \ell(x, {\Theta})}{\partial {x^i}}\| \in \text{top-k}].$
Formally, we define the approximation
\begingroup
\setlength\abovedisplayskip{-0.2mm}
\setlength\belowdisplayskip{-0.2mm}
\begin{equation}
   {\frac{\partial \ell(x, \Theta)}{\partial \Theta_{w}}}\vert_\text{top-k} = \sum_{i \in x^\text{top-k} ~\wedge~ x^i = w}\frac{\partial \ell(x, {\Theta})}{\partial {x^i}} 
\end{equation}
\endgroup
where $x^i$ is the word at position $i$, and $x^{top-k}$ is the set of top-k input positions by gradient norm. We then propose 
\begingroup
\setlength\abovedisplayskip{-1mm}
\setlength\belowdisplayskip{-4mm}
\begin{equation} \label{eq:trackemb-topk}
\begin{split}
    \text{TracIn-WE-Topk}(x,x) =  -\sum_{w \in  x \cap x'} \frac{\partial\ell(x, \Theta_{w})}{\partial \Theta_{w}}\vert_\text{top-k}^T \cdot \frac{\partial \ell(x', \Theta_{w})}{\partial \Theta_{w}}\vert_\text{top-k}.
\end{split}
\end{equation}
\endgroup
%Formally, $\frac{\partial \ell(x, \Theta)}{\partial \Theta_{w}}$ is approximated by $\sum_{x^i = w}\frac{\partial \ell(x, {\Theta})}{\partial {x^i}} \mathbbm{1}[\|\frac{\partial \ell(x, {\Theta})}{\partial {x^i}}\| \in \text{top-k}].$
%With this, we propose 
%\begin{equation*} \label{eq:trackemb-topk}
%\begin{split}\small
%    &\text{TracIn-WE-Topk}(x,x) = \\& -\sum_{w \in  x %\cap x'} \frac{\partial'\ell(x, \Theta_{w})}{\partial' %\Theta_{w}}^T \cdot \frac{\partial' \ell(x', %\Theta_{w})}{\partial' \Theta_{w}}.
%\end{split}
%\end{equation*}
\vspace{-1mm}
\paragraph{Computational complexity} Let $L$ be the max length of each sentence, $d$ be the word embedding dimension, and $o$ be the average overlap between two sentences. If the training and test point gradients are precomputed and saved then the average computation complexity for calculating TracIn-WE for $m$ training points and $n$ testing points is $O(m n  o  d)$. This can be contrasted with the average computation complexity for influence functions on the word embedding layer, which takes $O(mn d^2 v^2 + d^3 v^3)$, where $v$ is the vocabulary size which is typically larger than $10^4$, and $o$ is typically less than $5$. %Finally, we note that saving the word embedding gradients for TracIn-WE incurs space complexity of $O((m+n) L d)$, which may be costly when $m$ and $L$ are both very large (while influence function requires the space complexity $O((m+n) L d + d^2 v^2 $).
The approximation for TracIn-We-Topk drops 
%the space complexity from $O((m+n) L d_w)$ to $O((m+n)k d)$, and 
the computational complexity from $O(m  n  o  d)$ to $O(m n o_k d)$ where $o_k$ is the average overlap between the sets of top-k words from the two sentences. It has the additional benefit of preventing unimportant words (ones with small gradient) from dominating the word similarity by multiple occurrences, as such words may get pruned.
In all our experiments, we set $k$ to $10$ for consistency, and do not tune this hyper-parameter.
\vspace{-4mm}
\subsection{Influence without Word-Overlap}\label{sec:word_overlap}
\vspace{-3mm}
One potential criticism of  TracIn-WE is that it may not capture any influence when there are no overlapping words between $x$ and $x'$.
To address this, we note that modern NLP models often include a ``start'' and ``end'' token in all inputs.
We posit that gradients of the embedding weights of these tokens take into account the semantics of the input (as represented in the higher layers), and enable TracIn-WE to capture influence between examples that are semantically related but do not have any overlapping words.
%We argue that the word gradient similarity for these tokens can capture some form of influence based on semantic similarity even without word overlaps.
%However, this is not true due to the design of start and end token that is common to many NLP tasks. We suspect that the gradient similarity can capture some form of influence based on semantic similarity even without word overlaps. 
We illustrate this in S5-S7 in Tab.~\ref{tab:word_sim} via examples for the MNLI task. 
Sentence S5 has no overlapping words with S6 and S7. However, the word gradient similarity of ``start'' and ``end'' tokens for the pair S5-S6 is $1.15$, while that for the pair S5-S7 is much lower at $-0.05$.
Indeed, sentence S5 is more similar to S6 than S7 due to the presence of similar word pairs (e.g., think and thinks, annoying and boring), and the same negation usage. 
We further validate that TracIn-WE can capture influence from examples without word overlap via a controlled experiment in Sec.~\ref{sec:exp}.

%First, we observe that models based on the transformer architecture [CITE] featurize the input by including a ``start'' and ``end'' token.
%Thus, these tokens are common to all examples. 
\vspace{-2mm}

\begin{figure*}
    \centering
    \begin{minipage}{0.33\textwidth}
        \centering
        \includegraphics[width=1.0\textwidth]{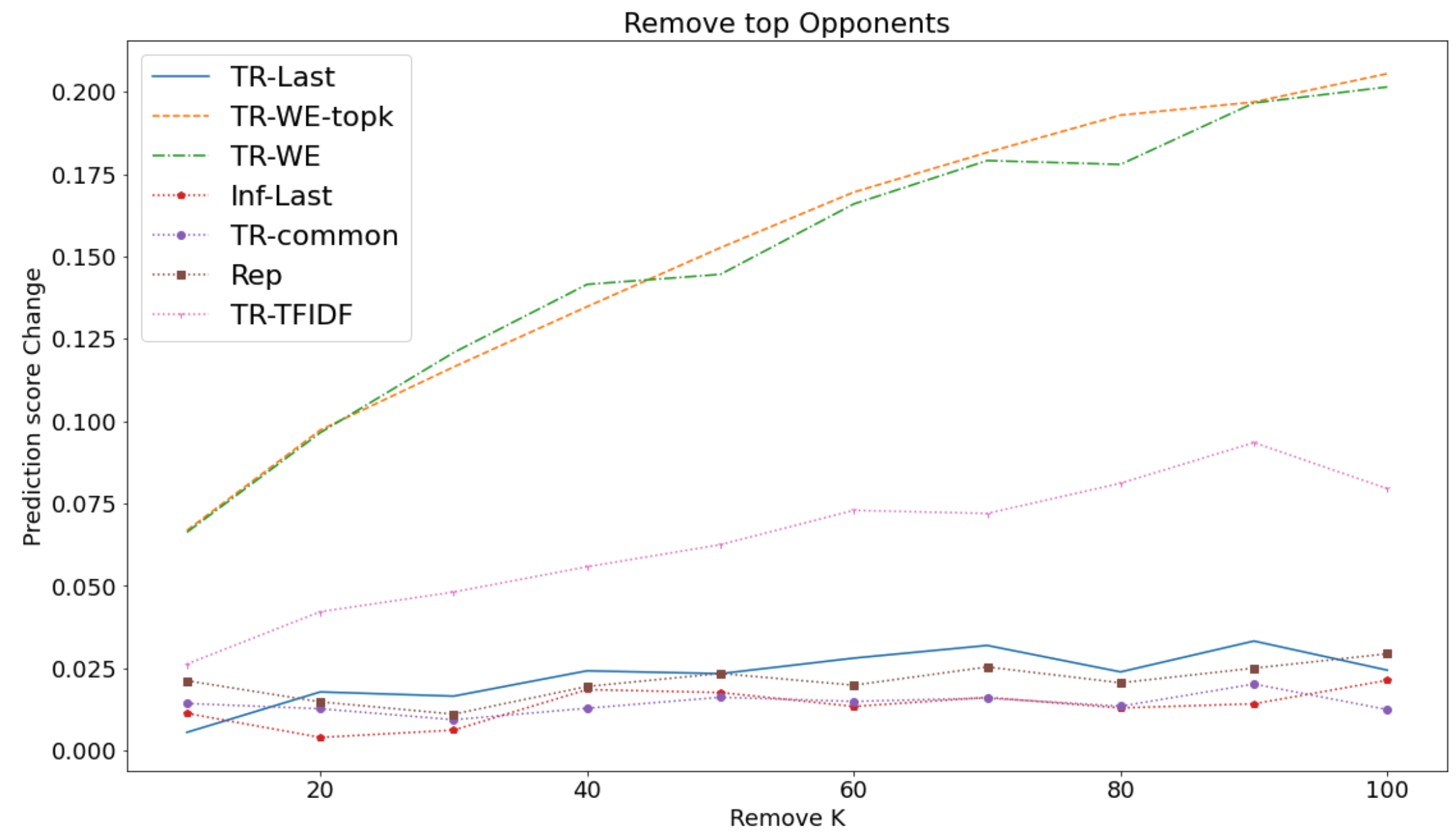} % first figure itself
    %\caption{Deletion curve for removing top opponents (larger is better).}
    \end{minipage}\hfill
    \begin{minipage}{0.33\textwidth}
        \centering
        \includegraphics[width=1.0\textwidth]{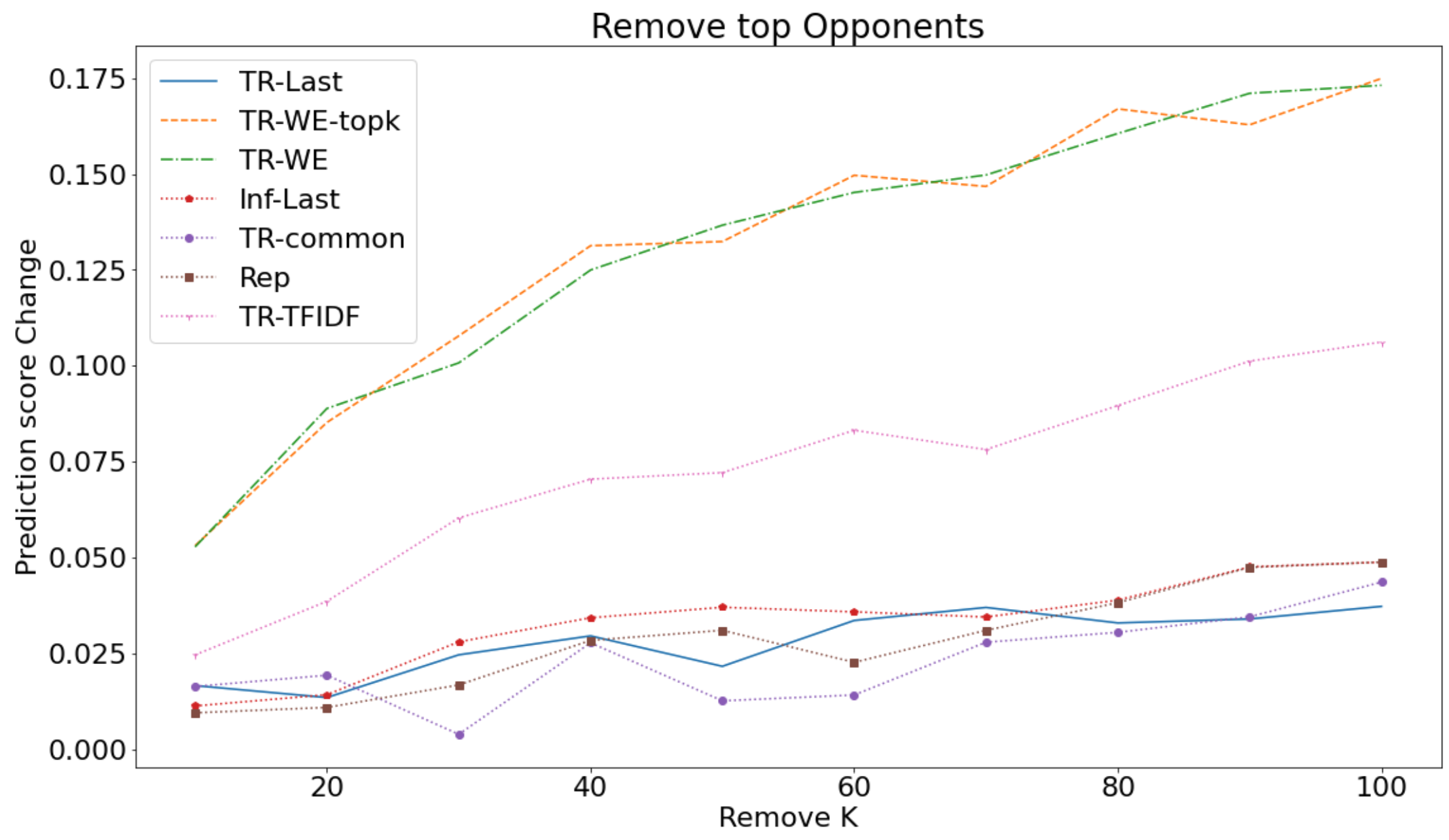} % second figure itself
        %\caption{Deletion curve for removing top proponents (smaller is better).}
    \end{minipage}
    \begin{minipage}{0.33\textwidth}
        \centering
        \includegraphics[width=1.0\textwidth]{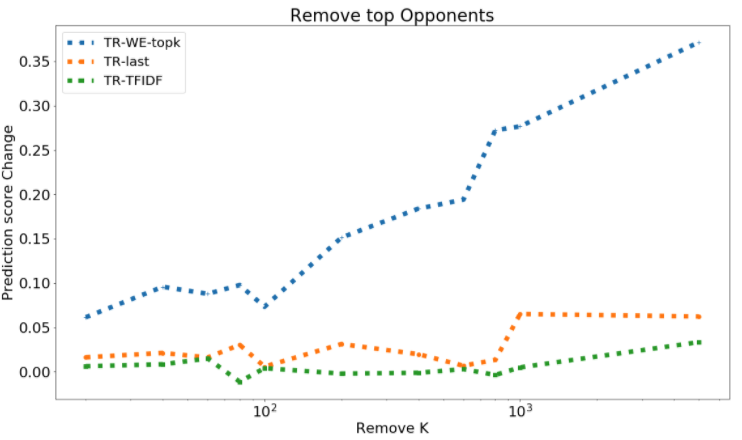} % second figure itself
        %\caption{Deletion curve for removing top proponents (smaller is better).}
    \end{minipage}
    \vspace{-4mm}
    %\caption{Deletion Curve for removing opponents (larger better) on Toxicity (left), AGnews (mid), and MNLI (right).}
    \label{fig:del_opp}
    \vspace{-3mm}
\end{figure*}\begin{figure*}
    \centering
    \begin{minipage}{0.33\textwidth}
        \centering
        \includegraphics[width=1.0\textwidth]{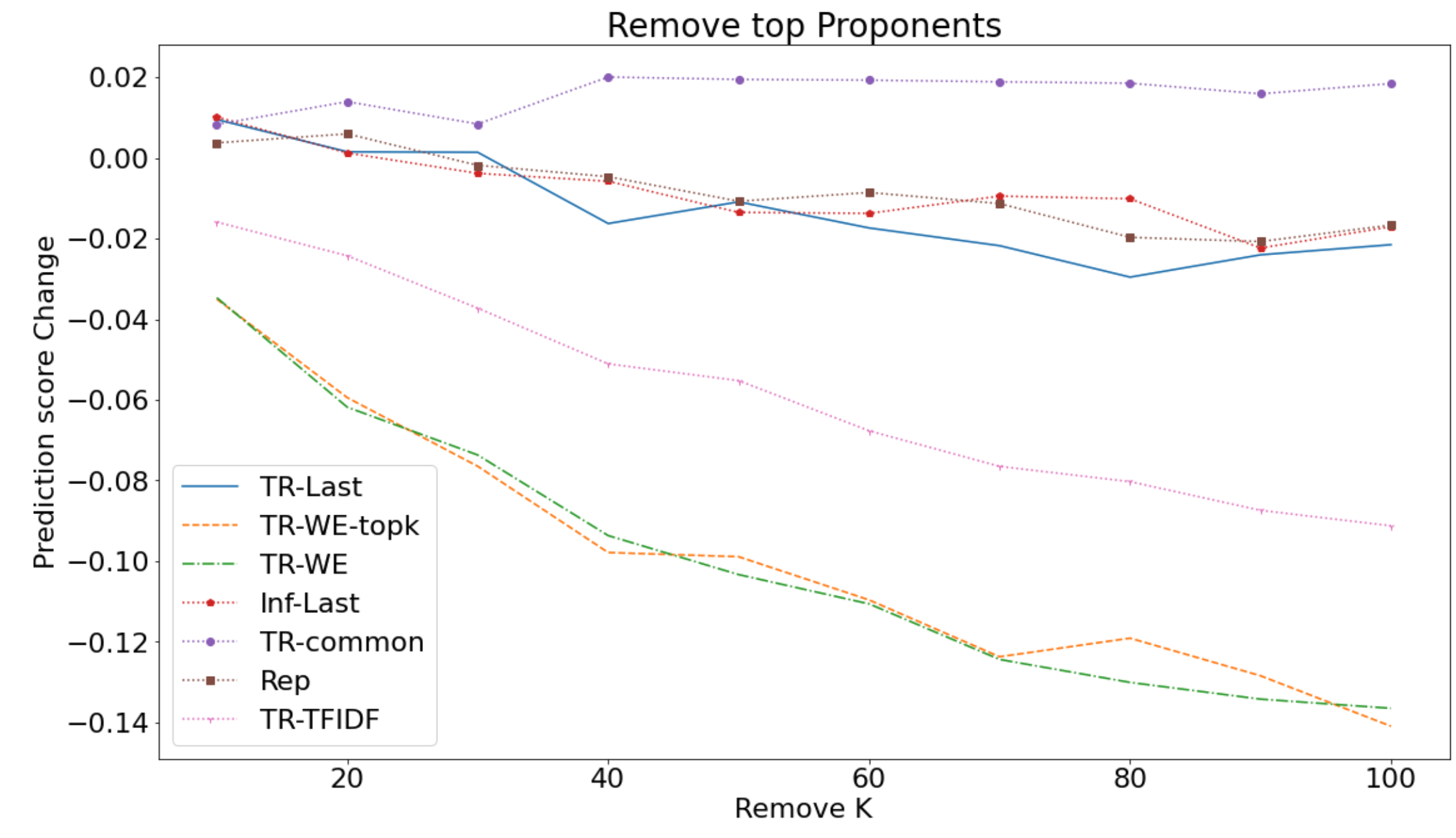} % first figure itself
    %\caption{Deletion curve for removing top opponents (larger is better).}
    \end{minipage}\hfill
    \begin{minipage}{0.33\textwidth}
        \centering
        \includegraphics[width=1.0\textwidth]{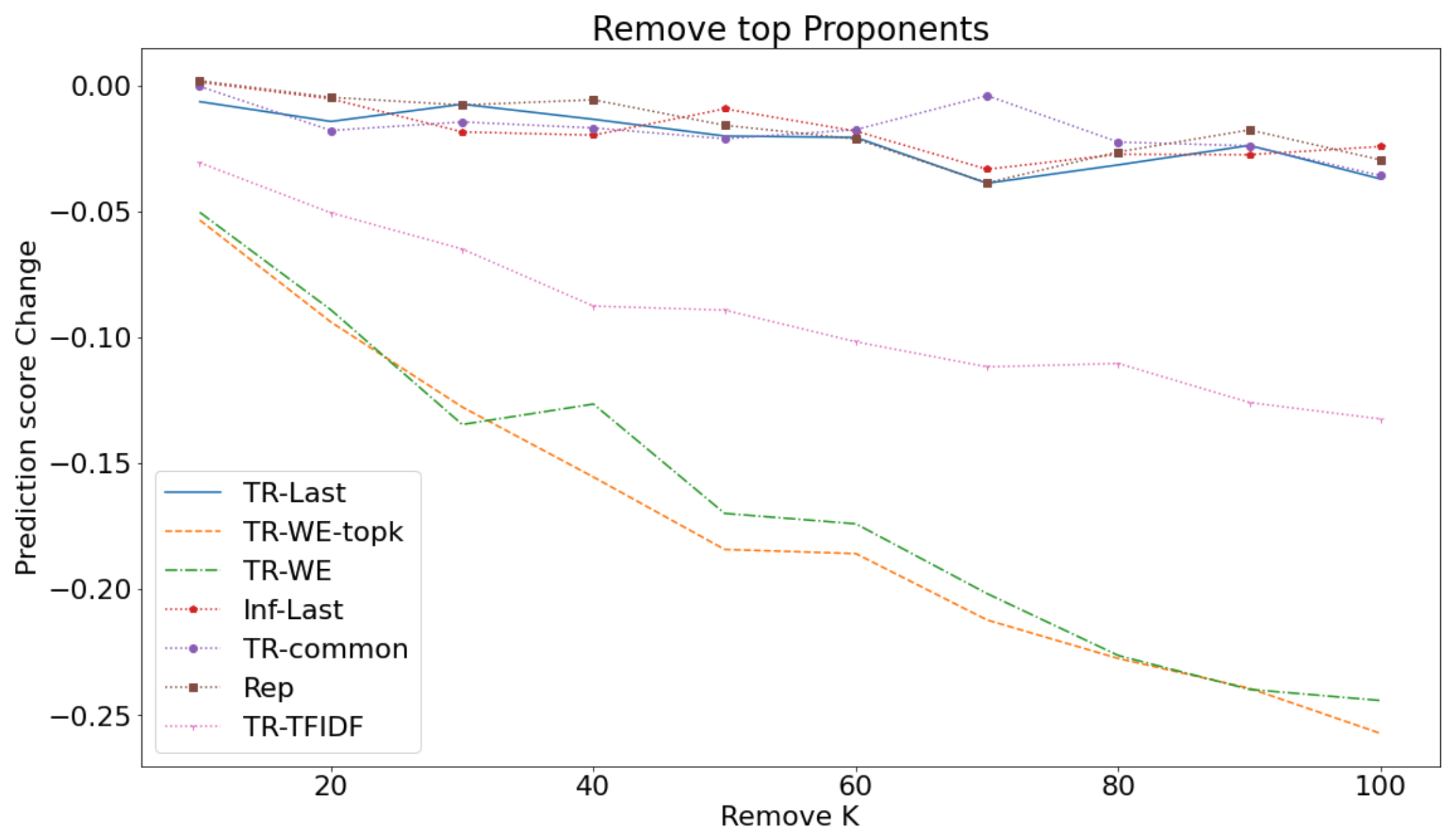} % second figure itself
        %\caption{Deletion curve for removing top proponents (smaller is better).}
    \end{minipage}
    \begin{minipage}{0.33\textwidth}
        \centering
        \includegraphics[width=1.0\textwidth]{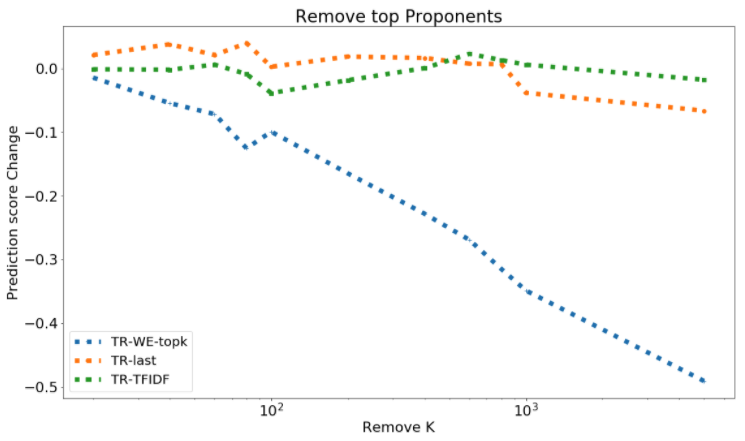} % second figure itself
        %\caption{Deletion curve for removing top proponents (smaller is better).}
    \end{minipage}
    \vspace{-4mm}
    \caption{Deletion Curve for removing opponents (top figure, larger better) and proponents (bottom figure, smaller better) on Toxicity (left), AGnews (mid), and MNLI (right).}
    \label{fig:del_pro}
    \vspace{-3mm}
\end{figure*}

 \begin{table*}[t!]
    \centering% This is an environment - we probably don't want the extra spacing of center in addition to that added by table etc.
    \caption{AUC-DEL table for various methods in different datasets. Highest number is bold.}
    \vspace{1mm}
    \label{tab:auc}
    \small
    \adjustbox{max width=0.98\columnwidth}{
    \begin{tabular}{p{0.08\textwidth} p{0.14\textwidth} p{0.09\textwidth}p{0.07\textwidth}p{0.07\textwidth}p{0.08\textwidth}p{0.12\textwidth}p{0.13\textwidth}p{0.12\textwidth}}
      \toprule% nicer rules courtesy of booktabs - but then we need to drop the verticals 
      Dataset & Metric & Inf-Last & Rep& TR-last& TR-WE& TR-WE-topk &  TR-TFIDF &  TR-common\\
      \midrule
      Toxic & AUC-DEL$+\downarrow$ & $-0.008$ & $-0.008$ & $-0.013$& $\mathbf{-0.100}$ & $-0.099$ & $-0.067$ &$\ \ \ 0.016$\\
      Bert & AUC-DEL$-\uparrow$ & $\ \ \ 0.014$ & $\ \ \ 0.021$ & $\ \ \ 0.023$& $\ \ \ 0.149$ & $\ \ \ \mathbf{0.151}$ & $\ \ \ 0.063$ & $\ \ \ 0.014$\\
      \midrule
      AGnews & AUC-DEL$+\downarrow$ & $-0.018$ & $-0.016$ & $-0.021$& $-0.166$ & $\mathbf{-0.174}$ & $-0.090$ & $-0.017$\\
      Bert & AUC-DEL$-\uparrow$ & $\ \ \ 0.033$ & $\ \ \ 0.028$ & $\ \ \ 0.028$& $\ \ \ {0.130}$ & $\ \ \ \mathbf{0.131}$ & $\ \ \  0.072$ & $\ \ \ 0.023$\\
      \midrule
      MNLI & AUC-DEL$+\downarrow$ & & & $\ \ \ 0.006$& & $\mathbf{-0.198}$ &  $-0.004$ \\
      Bert & AUC-DEL$-\uparrow$ & & & $ \ \ \ 0.026$& & $\ \ \ \mathbf{0.169}$ &  \ \ \ $0.005$ \\
      \midrule
      Toxic & AUC-DEL$+\downarrow$ & $-0.011$ & $-0.004$ &$ \ \ \ 0.001$ & ${-0.030}$ & $\mathbf{-0.038}$ & $-0.001$& $-0.001$ \\
      Roberta & AUC-DEL$-\uparrow$ & $ \ \ \ 0.023$ & $\ \ \ 0.012$ & $\ \ \ 0.003$ & $\ \ \ \mathbf{0.033}$ & $\ \ \ {0.030}$ & $\ \ \ 0.006$ & $\ \ \ 0.010$\\
      \midrule
       Dataset & Metric & Inf-Last & Rep& TR-last& TR-WE& TR-WE-topk & TR-WE-NoC & TR-common \\
       \midrule
        Toxic  & AUC-DEL$+\downarrow$ & $-0.009$ & $-0.008$ & $-0.006$& $\mathbf{-0.018}$ & $-0.016$ & $\ \ \ 0.003$ & $-0.008$ \\
        Nooverlap & AUC-DEL$-\uparrow$ & $ \ \ \ 0.008$ & $\ \ \ 0.007$ & $\ \ \ 0.010$& $\ \ \ \mathbf{0.026}$ & $\ \ \ \mathbf{0.026}$ & $\ \ \ 0.001$ & $\ \ \ 0.015$ \\
      \bottomrule
      %\vspace{-5mm}
    \end{tabular}}
    \vspace{-6mm}
  \end{table*}

\vspace{-4mm}
\section{Experiments} \label{sec:exp}
\vspace{-3mm}

We evaluate the proposed influence methods on $3$ different NLP classification datasets with BERT models. We choose a transformer-based model as it has shown great success on a series of down-stream tasks in NLP, and we choose BERT model as it is one of the most commonly used transformer model. For the smaller Toxicity and AGnews dataset, we operate on the Bert-Small model, as it already achieves good performance. For the larger MNLI dataset, we choose the Bert-Base model with $110M$ model parameters, which is a decently large model which we believe could represent the effectiveness of our proposed method on large-scale language models. As discussed in Section~\ref{sec:eval_metric}, we use the \emph{case deletion} evaluation and report the metrics on the deletion curve in Table \ref{tab:auc} for various methods and datasets. The standard deviation for all AUC values all methods is reported in Table \ref{tab:auc-std}.
%and an additional setting where there are not much overlaps between sentences.
\vspace{-3.5mm}
\paragraph{Baselines}
One question to ask is whether the good performance of TracIn-WE is a result that it captures the low-level word information well. To answer this question, we design a synthetic data influence score as the TF-IDF similarity~\cite{salton1988term} multiplied by the loss gradient dot product for $x$ and $x'$. TR-TFIDF can be understood by replacing the embedding similarity of TracIn-Last by the TF-IDF similarity, which captures low level similarity.
%\begin{equation} %\label{eq:TracIntfidf}
\begingroup
\begin{equation}
        \text{TR-TFIDF}(x,x') = - \text{Tf-Idf}(x,x')  \frac{\partial \ell(x, \Theta)}{\partial \fun(x, \Theta)} ^T \frac{\partial \ell(x', \Theta)}{\partial \fun(x, \Theta)}.
\end{equation}
\vspace{-2mm}
\endgroup

\vspace{-3mm}
%\end{equation}
\paragraph{Toxicity.}
We first experiment on the toxicity comment classification dataset \citep{toxic2018kaggle}, which contains sentences that are labeled toxic or non-toxic. We randomly choose $50,000$ training samples and $20,000$ validation samples. We then fine-tune a BERT-small model on our training set, which leads to $96 \%$ accuracy. Out of the $20,000$ validation samples, we randomly choose $20$ toxic and $20$ non-toxic samples, for a total of $40$ samples as our targeted test set.
%where our goal is to choose proponents and opponents based on each example in the test set. 
For each example $x'$ in the test set, we remove top-$k$ proponents and top-$k$ opponents in the training set respectively, and retrain the model to obtain $\textsc{del}_+(x', k, \I)$ and $\textsc{del}_-(x', k, \I)$ for each influence method $\I$. We vary $k$ over $\{10, 20, \hdots, 100\}$. 
For each $k$, we retrain the model $10$ times and take the average result, and then average over the $40$ test points. We implement the methods Influence-last, Representer Points, TracIn-last, TracIn-WE, TracIn-WE-Topk, TracIn-TFIDF (introduced in Sec.~\ref{sec:last-issue}), TracIn-common (which is a variant of TracIn only using the start token and end token to calculate gradient), and abbreviate TracIn with TR in the experiments. 
We see that our proposed TracIn-WE method, along with its variants TracIn-WE-Topk outperform other methods by a significant margin.
As mentioned in Sec.~\ref{sec:last}, TF-IDF based method beats the existing data influence methods using last layer weights by a decisive margin as well, but is still much worse compared to TracIn-WE. Therefore, TracIn-WE did not succeed by solely using low-level information. Also, we find that TracIn-WE performs much better than TracIn-common, which uses the start and end tokens only. This shows that the keyword overlaps (such as lazy, end in Table \ref{tab:decom}) is crucial to the great performance of TracIn-WE.
%This again showcases the issue of last-layer based influences being dominated by loss.

\vspace{-2mm}
\paragraph{AGnews.}
We next experiment on the AG-news-subset \citep{Agnews2015, zhang2015character}, which contains a corpus of news with $4$ different classes. We follow our setting in toxicity and choose $50,000$ training samples, $20,000$ validation samples, and fine-tune with the same BERT-small model that achieves $90\% $ accuracy on this dataset. We randomly choose $100$ samples with $25$ from each class as our targeted test set. The $\textsc{auc-del}_+$ and $\textsc{auc-del}_-$ scores for $k \in \{10, 20, \hdots, 100\}$ are reported in Table \ref{tab:auc}. 
Again, we see that the variants of TracIn-WE significantly outperform other existing methods applied on the last layer. 
In both AGnews and Toxicity, removing $10$ top-proponents or top-opponents for TracIn-WE has more impact on the test point compared to removing $100$ top-proponents or top-opponents for TracIn-last.
\vspace{-2mm}
\paragraph{MNLI.}
Finally, we test on a larger scale dataset, Multi-Genre Natural Language Inference (MultiNLI) ~\cite{N18-1101}, which consists of $433k$ sentence pairs with textual entailment information, including entailment, neutral, and contradiction. 
In this experiment, we use the full training and validation set, and BERT-base which achieves $84 \%$ accuracy on matched-MNLI validation set. 
We choose $30$ random samples with $10$ from each class as our targeted test set.
%to calculate $\textsc{del}_+(x',k,\I)$ and $\textsc{del}_-(x',k\,\I)$. 
We only evaluate TracIn-WE-Topk, TracIn-last and TracIn-TFIDF as those were the most efficient methods to run at large scale. We vary $k \in \{20, \hdots, 5000\}$, and the $\textsc{auc-del}_+$ and $\textsc{auc-del}_-$ scores for our test set are reported in Table \ref{tab:auc}.
%We only evaluate TR-last, TR-WE-topk, and TR-TFIDF to calculate $\textsc{del}_+(x')$ and $\textsc{del}_-(x')$ with $k \in \{20,40,60, 80,100, 200, 400, 600, 800, 1000\}$. 
%We choose TR-WE-topk as it is the most efficient version to run at large scale. 
Unlike previous datasets, here TracIn-TFIDF does not perform better than TracIn-Last, which may be because input similarity for MNLI cannot be merely captured by overlapping words. For instance, a single negation would completely change the label of the sentence.
However, we again see TracIn-WE-Topk significantly outperforms TracIn-Last and TracIn-TFIDF, demonstrating its efficacy over natural language understanding tasks as well.
This again provides evidence that TracIn-WE can capture both low-level information and high-level information. The deletion curve of Toxicity, AGnews, MNLI is in shown in Fig. \ref{fig:del_opp} and Fig. \ref{fig:del_pro}.
\vspace{-3mm}
\paragraph{Toxicity-Roberta.}
To additionally test whether our experiment results apply to more modern models, we repeat our experiments on the toxicity dataset with a Roberta model~\cite{liu2019roberta}, while fixing other settings. We find that the TracIn-WE and TracIn-WE-Topk still significantly outperforms other results. 

\vspace{-3mm}
\paragraph{No Word Overlap.}
To assess whether TracIn-WE can do well in settings where the training and test examples do not have overlapping words, we construct a controlled experiment on the Toxicity dataset. We follow all experimental setting for Toxicity classification with the Bert model, but making two additional changes --
%where only sentences with very little word overlap is allowed. 
(1) given a test sentence $x'$, we only consider the top-$5000$ training sentences (out of $50,000$) with the least word overlap for computing influence. 
We use TF-IDF similarity to rank the number of word overlaps so that stop word overlap will not be over-weighted.
(2) We also fix the token embedding during training (result when word-embedding is not fixed is in the appendix, where removing examples based on any influence method does not change the prediction), as we find sentence with no word overlaps carry more influence when the token embedding is fixed. %Thus, we use the same data and model and $k$ as in our previous experiment on Toxicity, where the only difference is the word-embedding layer is freezed during training, and only examples with the least overlap is allowed to be removed. 
The $\textsc{auc-del}_+$ and $\textsc{auc-del}_-$ scores are reported in the lower section of Table \ref{tab:auc}. 
We find that TracIn-WE variants can outperform last-layer based influence methods even in this controlled setting, showing that TracIn-WE can retrieve influential examples even without non-trivial word overlaps.
%without replying on word overlaps.
In Section~\ref{sec:word_overlap}, we claimed that this gain stems from the presence of common tokens (``start'', ``end'', and other frequent words). 
To validate this, we compared with a controlled variant, TracIn-WE-NoCommon (TR-WE-NoC) where the common tokens are removed from TracIn-WE.
As expected, this variant performed much worse on the $\textsc{auc-del}_+$ and $\textsc{auc-del}_-$scores, thus confirming our claim.
We also find that the result of TracIn-WE is better than TracIn-common (which is TracIn-WE with only ``start'' and ``end'' tokens), which shows that the common tokens such as stop words and punctuation may also help finding influential examples without meaningful word overlaps.

%that the success of TracIn-WE in this setting stems from the common tokens.
%can be attributed to the existence of common tokens when there are very few word overlaps in the training set.

\vspace{-4mm}
\section{Related Work}
\vspace{-3mm}
In the field of explainable machine learning, our works belongs to training data importance \citep{koh2017understanding, yeh2018representer, jia2019towards, pruthi2020estimating, khanna2018interpreting, sui2021representer}. Other forms of explanations include feature importance feature-based explanations, gradient-based explanations \citep{baehrens2010explain,simonyan2013deep,zeiler2014visualizing,bach2015pixel,ancona2017unified, sundararajan2017axiomatic, shrikumar2017learning, ribeiro2016should,lundberg2017unified, yeh2019on, petsiuk2018rise} and perturbation-based explanations \citep{ribeiro2016should,lundberg2017unified, yeh2019on, petsiuk2018rise}, self-explaining models \citep{wang2015falling, lee2019functional, chen2019looks}, counterfactuals to change the outcome of the model \citep{Wachter2017Counterfactual,dhurandhar2018explanations,Hendricks2018GroundingVE,vanderwaa2018, goyal2019counterfactual}, concepts of the model \citep{kim2018interpretability, zhou2018interpretable}. For applications on applying data importance methods on NLP tasks, there have been works identifying data artifacts~\citep{han2020explaining, pezeshkpour2021combining} and improving models~\citep{han2020fortifying, han2021influence} based on existing data importance method using the influence function or TracIn. In this work, we discussed weight parameter selection to reduce cancellation effect for training data attribution. There has been works that discuss how to cope with cancellation in the context of feature attribution: \citet{liu2020penalty} discusses how regularization during training reduces cancellation of feature attribution, \citet{kapishnikov2021guided} discusses how to optimize IG paths to minimize cancellation of IG attribution, and \citet{sundararajan2019exploring} discusses improved visualizations to adjust for cancellation.

\vspace{-4mm}
\section{Conclusion}
\vspace{-3mm}
In this work, we revisit the common practice of computing training data influence using only last layer parameters.
%In this work, we challenge the usage of last layer parameters in calculating data influence.
We show that last layer representations in language classification models can suffer from the cancellation effect, which in turn leads to inferior results on influence. 
%We first formulate the deletion curve evaluation, and show that the reductive last-layer similarity leads to poor result of data influence calculated based on the last layer parameters. 
We instead recommend computing influence on the word embedding parameters, and apply this idea to propose a variant of TracIn called TracIn-WE.
We show that TracIn-WE significantly outperforms last versions of existing influence methods on three different language classification tasks for several models, and also affords a word-level decomposition of influence that aids interpretability.
%We instead  propose the method TracIn-WE, using the word embedding parameters, and surprisingly show that the resulting influence method can capture high-lever information, and performs significantly better than other methods on the deletion curve evaluation. 

\clearpage
\bibliography{references}
\bibliographystyle{icml2022}
\section*{Checklist}

\iffalse

%%% BEGIN INSTRUCTIONS %%%
The checklist follows the references.  Please
read the checklist guidelines carefully for information on how to answer these
questions.  For each question, change the default \answerTODO{} to \answerYes{},
\answerNo{}, or \answerNA{}.  You are strongly encouraged to include a {\bf
justification to your answer}, either by referencing the appropriate section of
your paper or providing a brief inline description.  For example:
\begin{itemize}
  \item Did you include the license to the code and datasets? \answerYes{See Section~\ref{gen_inst}.}
  \item Did you include the license to the code and datasets? \answerNo{The code and the data are proprietary.}
  \item Did you include the license to the code and datasets? \answerNA{}
\end{itemize}
Please do not modify the questions and only use the provided macros for your answers.  Note that the Checklist section does not count towards the page limit.  In your paper, please delete this instructions block and only keep the Checklist section heading above along with the questions/answers below.
%%% END INSTRUCTIONS %%%
\fi

\begin{enumerate}

\item For all authors...
\begin{enumerate}
  \item Do the main claims made in the abstract and introduction accurately reflect the paper's contributions and scope?
    \answerYes{}
  \item Did you describe the limitations of your work?
    \answerYes{In Sec. \ref{sec:limit}}
  \item Did you discuss any potential negative societal impacts of your work?
     \answerYes{In Sec. \ref{sec:social}}
  \item Have you read the ethics review guidelines and ensured that your paper conforms to them?
    \answerYes{}
\end{enumerate}

\item If you are including theoretical results...
\begin{enumerate}
  \item Did you state the full set of assumptions of all theoretical results?
    \answerNA{}
        \item Did you include complete proofs of all theoretical results?
    \answerNA{}
\end{enumerate}

\item If you ran experiments...
\begin{enumerate}
  \item Did you include the code, data, and instructions needed to reproduce the main experimental results (either in the supplemental material or as a URL)?
    \answerNo{}
  \item Did you specify all the training details (e.g., data splits, hyperparameters, how they were chosen)?
    \answerYes{}
        \item Did you report error bars (e.g., with respect to the random seed after running experiments multiple times)?
    \answerYes{}
        \item Did you include the total amount of compute and the type of resources used (e.g., type of GPUs, internal cluster, or cloud provider)?
    \answerYes{In Sec. \ref{sec:compute}}
\end{enumerate}

\item If you are using existing assets (e.g., code, data, models) or curating/releasing new assets...
\begin{enumerate}
  \item If your work uses existing assets, did you cite the creators?
    \answerYes{}
  \item Did you mention the license of the assets?
    \answerYes{In Sec. \ref{sec:lincence}}
  \item Did you include any new assets either in the supplemental material or as a URL?
    \answerNo{}
  \item Did you discuss whether and how consent was obtained from people whose data you're using/curating?
    \answerNA{Only used public data.}
  \item Did you discuss whether the data you are using/curating contains personally identifiable information or offensive content?
    \answerNA{}
\end{enumerate}

\item If you used crowdsourcing or conducted research with human subjects...
\begin{enumerate}
  \item Did you include the full text of instructions given to participants and screenshots, if applicable?
    \answerNA{}
  \item Did you describe any potential participant risks, with links to Institutional Review Board (IRB) approvals, if applicable?
    \answerNA{}
  \item Did you include the estimated hourly wage paid to participants and the total amount spent on participant compensation?
    \answerNA{}
\end{enumerate}

\end{enumerate}

\clearpage
\appendix

\section{Finding Mislabeling Patterns by Clustering} \label{sec:cluster}

We first select training data that are classified incorrectly at least $40\%$ during training (with early stopping) after training $20$ models in the AGnews dataset, which are more likely to consist of mislabeling examples. Our goal is to find mislabeled training examples that may consist similar mislabeling patterns. We hypothesize that two mislabeled training examples that have high influence to each other may be more likely to have a similar mislabeling patterns. Thus, we define the influence distance between training data as the negative of a scaled data influence.
\[\text{Inf-dis}_{\I}(x_a, x_b) = \max(1-\frac{\I(x_a,x_b)}{\sqrt{\I(x_a,x_a) \I(x_b,x_b)}} , 0)\]
\vspace{-2mm}

Thus, if data B is a strong proponent to data A, the influence distance would be small. We then cluster these ``difficult to learn” examples by the influence distance, and we apply this clustering on AGnews, where the influence distance is calculated by TracIn-WE on a CNN model, and use Agglomerative Clustering with threshold $0.8$, which results in $31$ clusters with at least $3$ elements.

\begin{table*}[!h]
    \vspace{-3mm}
    \centering% This is an environment - we probably don't want the extra spacing of center in addition to that added by table etc.
    \caption{Descriptions of Clusters for Clustering mislabeling examples}
    \label{tab:mistake_cluster_descriptions}
    \small
    \adjustbox{max width=1.0\columnwidth}{
    \begin{tabular}{p{0.08\textwidth} p{0.45\textwidth}p{0.2\textwidth} p{0.12\textwidth}p{0.1\textwidth}}
      \toprule
      & Cluster Information & Common Words&Predict Label& True Label \\
      \toprule% nicer rules courtesy of booktabs - but then we need to drop the verticals 
       Cluster 1 &  Red Sox and Yankees AL championship series& championship, series & World & Sport \\
      \midrule
      Cluster 2 &   The same/similar sentence repeats 12 times&  has, focus, priority  & Tech/Science & Tech/Science \\
      \midrule
      Cluster 3 &  The same/similar sentence repeats 10 times& priority, fourth  & Business & Tech/Science \\
      \midrule
      Cluster 4 &  Oracle’s takeover for PeopleSoft&  peoples, \#\#oft  & Tech/Science & Business \\
      \midrule
      Cluster 5 &  Ryder Cup &ryder, cup & World& Sport \\
      \midrule
      \bottomrule
    \end{tabular}
    }
    \vspace{-6mm}
  \end{table*}
  
  \begin{table*}[!h]
    \centering% This is an environment - we probably don't want the extra spacing of center in addition to that added by table etc.
    \caption{Sentences in Clusters for Clustering mislabeling examples}
    \label{tab:mistake_cluster_sentences}
    \small
    \adjustbox{max width=1.0\columnwidth}{
    \begin{tabular}{p{0.08\textwidth} p{0.9\textwidth} }
      \toprule
      Cluster & Cluster Examples \\
      \midrule% nicer rules courtesy of booktabs - but then we need to drop the verticals 
       Cluster 1 &  -- BOSTON - The New York Yankees and Boston were tied 4-4 after 13 innings Monday night with the Red Sox trying to stay alive in the AL championship series. \\
        &  -- Steady rain Friday night forced major league baseball to postpone Game 3 of the AL championship series between the Boston Red Sox and New York Yankees. \\
        &  -- After Curt Schilling and Pedro Martinez failed to get the Boston Red Sox a win against the New York Yankees in the first two games of the AL championship series \\
        &  -- The Boston Red Sox entered this AL championship series hoping to finally overcome their bitter rivals from New York following a heartbreaking seven-game defeat last October. \\
      \midrule
       Cluster 2 & -- com October 13, 2004, 5:06 PM PT. This fourth priority‘s main focus has been enterprise directories as organizations spawn projects around identity infrastructure. \\
        &  -- com September 15, 2004, 11:03 AM PT. This fourth priority’s main focus has been improving or obtaining CRM and ERP software for the past year and a half. \\
        &  -- com October 11, 2004, 11:16 AM PT. This fourth priority’s main focus has been enterprise directories as organizations spawn projects around identity infrastructure. \\
      \midrule
       Cluster 3 &  -- This fourth priority’s main focus has been enterprise directories as organizations spawn projects around identity infrastructure. \\
        &  -- com September 13, 2004, 8:58 AM PT. This fourth priority’s main focus has been improving or obtaining CRM and ERP software for the past year and a half. \\
        &  -- com October 26, 2004, 7:41 AM PT. This fourth priority’s main focus has been enterprise directories as organizations spawn projects around identity infrastructure. \\
      \midrule
       Cluster 4 &  -- The Wall Street rumor mill is working overtime, spinning off speculation about how soon a decision will be announced in the US government lawsuit aimed at blocking Oracle’s proposed takeover of PeopleSoft. \\
        &  -- Oracle Corp. has extended its \$7.7 billion hostile takeover bid for Pleasanton’ PeopleSoft Inc. until Sept. 10. Redwood City-based Oracle’s previous offer would have expired at 9 pm Friday. \\
        &  -- A director of the Oracle Corporation testified that the company's \$7.7 billion hostile bid for PeopleSoft might not be the final offer. \\
      \midrule
       Cluster 5 &  -- BLOOMFIELD TOWNSHIP, Mich. - For the first time in three days at the Ryder Cup, there was plenty of red on the scoreboard - as in American red, white and blue... \\
        &  -- Europe go into the singles needing three-and-a-half points to win the Ryder Cup. \\
        &   -- BLOOMFIELD TOWNSHIP, Mich. - Staring down Tiger Woods, Phil Mickelson and the rest of the Americans, Europe got off to a stunning start Friday in the Ryder Cup... \\
      \midrule
      \bottomrule
    \end{tabular}
    }
    \vspace{-3mm}
  \end{table*}

We show $5$ clusters where we find that the examples in the clusters are clear mispredictions, and report the description of clusters in Tab. \ref{tab:mistake_cluster_descriptions} and the actual sentences (sometimes abbreviated) in Tab. \ref{tab:mistake_cluster_sentences}. We report the most common words in clusters by recording the top-$5$ words that contributed to the TracIn-WE score for each pair of sentences, and report the words that are top-$5$ in most pair of sentences in the clusters. We note that while cluster 2 are not necessary mislabels, the combination with cluster 3 shows some repetition and inconsistency issues of the labeling process of Agnews. Other clusters clearly demonstrate very clear mislabel patterns in AGnews, which can be fixed systemically by humans writing a fixing function, which can be an interesting follow-up direction.

\section{Limitations of Our Work} \label{sec:limit}
While we believe that our claim that ``first is better than last for training data influence'' is general, we did not test out the method on all data modalities and all types of models, as the computation of deletion score is very expensive. We note that we have not tested TracIn-Last for generative tasks, as it is beyond the scope of the paper, and we leave it to future works.

\section{Potential Social Impact of Our Work} \label{sec:social}
One potential social impact is that one may use the algorithm to adjust training data to effect a particular test point's prediction. This can be used for good (making the model more fair), or for bad (making the model more biased).

\section{Computation} \label{sec:compute}
We report the run time for TracIn-WE, TracIn-Sec, TracIn-Last, Inf-Sec, Inf-Last for the CNN text model. The second convolutional layer has $6400$ parameters, last layer has $440$ parameters, token embedding layer has $3.2$ million parameters. We applied these methods on $50000$ training points and $10$ test points. The preprocessing time (sec) per training point is  $0.004$, $0.004$, $0.003$, $3.52$, $0.002$, and the cost of computing influence per training point and test point pair (sec) is: $4 \cdot 10^{-4}$, $3 \cdot 10^{-5}$, $8 \cdot 10^{-6}$, $10^{-1}$, $2\cdot10^{-5}$. 
Influence function on the second layer is already order of magnitudes slower than other variants, and cannot scale to the word embedding layer with millions of parameters.

For remove and retrain on Toxicity and AGnews, we run our experiments on multiple V100 clusters. For remove and retrain on MNLI, we run our experiments on multiple TPU-v3 clusters. For toxicity and AGnews experiment, we need to fine-tune the language model on the classification task for $40 \times 6 \times 10 \times 2 \times 10$ times, where the fine-tuning takes around $10-20$ GPU-minute on a V100 for Bert-Small, and $40, 6, 10, 2, 10$ stands for number of test points, number of methods, removal numbers, proponents/ opponents, and repetition numbers respectively. On MNLI, we fine-tuned the language model for $19800 (30 \times 3 \times 11 \times 2 \times 10)$ times, where fine-tuning MNLI on BERT-Base takes around $320$ TPU-minute on a TPU-v3 cluster for Bert-base.

\section{Licence of Datatset} \label{sec:lincence}

Toxicity dataset has license cc0-1.0, AGnews dataset has license non-commercial use, and MNLI has license cc-by-3.0.

\section{Standard Deviation of Experiments}
We report the standard deviation of all AUC-DEL value reported in Tab.\ref{tab:auc}. To calculate each AUC-DEL score, we take the average after retraining $10$ times. We could then measure the standard deviation by bootstrapping. We report the number in Tab. \ref{tab:auc-std}, and we see that all methods have similar standard deviations.

 \begin{table*}[h!]
    \centering% This is an environment - we probably don't want the extra spacing of center in addition to that added by table etc.
    \caption{standard deviations for AUC-DEL table for various methods.}
    \vspace{1mm}
    \label{tab:auc-std}
    \small
    \adjustbox{max width=0.98\columnwidth}{
    \begin{tabular}{p{0.08\textwidth} p{0.14\textwidth} p{0.09\textwidth}p{0.07\textwidth}p{0.07\textwidth}p{0.08\textwidth}p{0.12\textwidth}p{0.13\textwidth}p{0.12\textwidth}}
      \toprule% nicer rules courtesy of booktabs - but then we need to drop the verticals 
      Dataset & Metric & Inf-Last & Rep& TR-last& TR-WE& TR-WE-topk &  TR-TFIDF &  TR-common\\
      \midrule
      Toxic & AUC-DEL$+\downarrow$ & $\ \ \ 0.004$ & $\ \ \ 0.006$ & $\ \ \ 0.004$& $\ \ \ 0.003$ & $\ \ \ 0.003$ & $\ \ \ 0.003$ &$\ \ \ 0.004$\\
      Bert & AUC-DEL$-\uparrow$ & $\ \ \ 0.004$ & $\ \ \ 0.003$ & $\ \ \ 0.003$& $\ \ \ 0.004$ & $\ \ \ 0.004$ & $\ \ \ 0.005$ & $\ \ \ 0.002$\\
      \midrule
      AGnews & AUC-DEL$+\downarrow$  & $\ \ \ 0.003$ & $\ \ \ 0.006$ & $\ \ \ 0.006$& $\ \ \ 0.007$ & $\ \ \ 0.005$ & $\ \ \ 0.005$ &$\ \ \ 0.004$\\
      Bert & AUC-DEL$-\uparrow$ & $\ \ \ 0.006$ & $\ \ \ 0.007$ & $\ \ \ 0.005$& $\ \ \ 0.006$ & $\ \ \ 0.003$ & $\ \ \ 0.005$ & $\ \ \ 0.003$\\
      \midrule
      MNLI & AUC-DEL$+\downarrow$ & & & $\ \ \ 0.017$& & $\ \ \ 0.014$ &  $\ \ \ 0.020$ \\
      Bert & AUC-DEL$-\uparrow$ & & & $ \ \ \ 0.026$& & $\ \ \ 0.014$ &  \ \ \ $0.010$ \\
      \midrule
      Toxic & AUC-DEL$+\downarrow$  & $\ \ \ 0.021$ & $\ \ \ 0.027$ & $\ \ \ 0.014$& $\ \ \ 0.015$ & $\ \ \ 0.013$ & $\ \ \ 0.020$ &$\ \ \ 0.022$\\
      Roberta & AUC-DEL$-\uparrow$  & $\ \ \ 0.019$ & $\ \ \ 0.019$ & $\ \ \ 0.022$& $\ \ \ 0.019$ & $\ \ \ 0.020$ & $\ \ \ 0.021$ &$\ \ \ 0.016$\\
      \midrule
       Dataset & Metric & Inf-Last & Rep& TR-last& TR-WE& TR-WE-topk & TR-WE-NoC & TR-common \\
       \midrule
        Toxic  & AUC-DEL$+\downarrow$  & $\ \ \ 0.004$ & $\ \ \ 0.004$ & $\ \ \ 0.003$& $\ \ \ 0.005$ & $\ \ \ 0.003$ & $\ \ \ 0.004$ &$\ \ \ 0.003$\\
        Nooverlap & AUC-DEL$-\uparrow$  & $\ \ \ 0.004$ & $\ \ \ 0.003$ & $\ \ \ 0.003$& $\ \ \ 0.004$ & $\ \ \ 0.004$ & $\ \ \ 0.004$ &$\ \ \ 0.003$\\
      \bottomrule
      %\vspace{-5mm}
    \end{tabular}}
    \vspace{-4mm}
  \end{table*}

\section{A different viewpoint on Issues with Last Layer.} \label{sec:last-issue}

We present our analysis in the context of the TracIn method applied to the last layer, referred to as TracIn-Last, although our experiments in Section \ref{sec:exp} suggest that Influence-Last and Representer-Last may also suffer from similar shortcomings.
For TracIn-Last, the similarity term $S(x,x') =  \nabla_{\Theta} \fun(x, \Theta_{last})^T \nabla_{\Theta} \fun(x', \Theta_{last})$ becomes $\act(x,\Theta_{last})^T \act(x',\Theta_{last})$ where $\act(x,\Theta_{last})$ is the final activation layer. We refer to it as \emph{last layer similarity}. 
Overall,
%When applied on the last layer, 
TracIn-last has the following formultation:
\vspace{-2mm}
\begin{equation*} \label{eq:TracInlast}
    \text{TracIn-Last}(x,x') = \act(x,\Theta)^T \act(x',\Theta) \frac{\partial \ell(x', \Theta)}{\partial \fun(x, \Theta)} ^T \frac{\partial \ell(x', \Theta)}{\partial \fun(x, \Theta)}.
\end{equation*}

\vspace{-2mm}

%Other top proponents and top opponents for the test example in Table \ref{tab:last_exp} were also found to be in top influential examples for other test examples.
%We now argue that TracIn-Last leads to inferior results and attribute this to shortcomings of last layer similarity.
We begin by qualitatively analyzing the influential examples from TracIn-Last, and find the top proponents to be unrelated to the test example.
We also observe that the top proponents of different test examples coincide a lot; see appendix ~\ref{sec:last-issue} for details.
This leads us to suspect that the top influence scores from TracIn-Last are dominated by the loss salience term of the training point $x$ (which is independent of $x'$), and not as much by the similarity term, which is also observed by \citet{barshan2020relatif, hanawa2021evaluation}.
Indeed, we find that on the toxicity dataset, the top-100 examples ranked by TracIn-Last and the top-100 examples ranked by the loss salience term $\frac{\partial \ell(x, \Theta)}{\partial \fun(x, \Theta)}$ have $49$ overlaps on average, while the top-100 examples by TracIn-Last and the top-100 examples ranked by the similarity term $\act(x,\Theta)^T \act(x',\Theta)$ have only $22$ overlaps on average. 
Finally, we find that replacing the last-layer similarity component by the well-known TF-IDF significantly improves its performance on the case deletion evaluation.
In fact, this new method, which we call TracIn-TDIDF, also outperforms Influence-Last, and Representer-Last on the case deletion evaluation; see Section~\ref{sec:exp} and Appendix ~\ref{sec:last-issue}.
We end this section with the following hypothesis.

\begin{hypothesis}
TracIn-Last and other influence methods that rely on last layer similarity fail in finding influential examples since last layer representations are too reductive and do not offer a meaningful notion of sentence similarity that is essential for influence.
%As a result, the ranking of such influence methods is dominated by the loss of the training examples.
\end{hypothesis}

We begin by qualitatively examining the influential examples obtained from TracIn-Last.
Consider the test sentence and its top-2 proponents and opponents in Table \ref{tab:last_exp}.
As expected, the proponents have the same label as the test sentence.
However, besides this label agreement, it is not clear in what sense the proponents are similar to the test sentence.
We also observe that out of $40$ randomly chosen test examples, proponent-1 is either in the top-20 proponents or top-20 opponents for $39$ test points.

 \begin{table*}[!hbt]
    \centering% This is an environment - we probably don't want the extra spacing of center in addition to that added by table etc.
    \caption{Examples for TracIn-Last }
    \label{tab:last_exp}
    \small
    \begin{tabular}{p{0.12\textwidth} p{0.70\textwidth} p{0.1\textwidth}}
      \toprule
      & Sentence content & Label \\
      \toprule% nicer rules courtesy of booktabs - but then we need to drop the verticals 
      Test Sentence & Somebody that double clicks your nick should have enough info but don’t let that cloud your judgement! There are other people you can hate for no reasons whatsoever. Hate another day. & Non-Toxic  \\\midrule% Note that there is no & before the first column - & only comes between columns so if you define n columns, you can have at most n-1 & symbols in any row
      Proponent-1 & Wow! You really are a piece of work, aren't you pal? Every time you are proven wrong, you delete the remarks. You act as though you have power, when you really don't. & Non-Toxic.\\% No need for 6 rows of space for each title!
      Proponent-2 & Ok i am NOT trying to piss you off ,but dont you find that touching another women is slightly disgusting. with all due respect,  dogblue & Non-Toxic\\% No need for 6 rows of space for each title!
      \midrule
      Opponent-1 & Spot, grow up!  The article is being improved with the new structure.  Please stop your nonsense. & Toxic \\% No need for 6 rows of space for each
      Opponent-2 & are you really such a cunt? (I apologize in advance for certain individuals who are too sensitive) & Toxic \\% No need for 6 rows of space for each title!
      \bottomrule
    \end{tabular}
    \vspace{-3mm}
  \end{table*}

To further validate that the inferior results from TracIn-Last can be attributed to the use of last layer similarity, we perform a controlled experiment where we replace the similarity term by a common sentence similarity measure --- the TF-IDF similarity~\cite{salton1988term}.
%\begin{equation} %\label{eq:TracIntfidf}
\[
    \text{TR-TFIDF}(x,x') = - \text{Tf-Idf}(x,x')  \frac{\partial \ell(x, \Theta)}{\partial \fun(x, \Theta)} ^T \frac{\partial \ell(x', \Theta)}{\partial \fun(x, \Theta)}
    \]
%\end{equation}
We find that $\text{TFIDF}$ performs much better than TracInCP-last and Influence-Last on the Del+ and Del- curve (see Fig. \ref{fig:del_pro}.
This shows that last layer similarity does not provide a useful measure of sentence similarity for influence.
 
%However, TF-IDF similarity is model-agnostic, and it is natural to ask whether we can replace TF-IDF similarity by some sentence similarity that is related to the model of interest while containing sufficient the low-level similarity.

%We also simply compare with the naive TF-IDF with label matching. The intuition of label matching is that examples with the same labels are usually proponents to each other, and examples with the different labels are usually opponents to each other. We thus multiply the tf-idf similarity by the label matching function.
%\begin{equation} %\label{eq:simpletfidf}
%    \I_{\text{LM-TFIDF}}(x,x') = - %\text{Tf-Idf}(x,x') %(\mathbbm{1}[y=y'] - %\mathbbm{1}[y\not=y']),
%\end{equation}
%where y is the label to data $x$ and $y'$ is the label to data $x'$. 
%We find that both $\I_{\text{TFIDF}}$ and $\I_{\text{TR-TFIDF}}$ performs much better than TracInCP-last and Influence-Last on the Del+ and Del- curve (see Fig. \ref{fig:del_agnews}, Fig. \ref{fig:del_toxic}). This is evidence that the embedding similarity did not provide useful instance similarity from the aspect of data influence. Since TF-IDF similarity captures sentence similarity in the form of low-level features, we hypothesis that the embedding similarity may no longer contain low-level similarity information, which could be useful for data influence. However, TF-IDF similarity is model-agnostic, and it is natural to ask whether we can replace TF-IDF similarity by some sentence similarity that is related to the model of interest while containing sufficient the low-level similarity.
Since TF-IDF similarity captures sentence similarity in the form of low-level features (i.e., input words), we speculate that last layer representations are too reductive and do not preserve adequate low-level information about the input, which is useful for data influence.
This is aligned with existing findings that  last layer similarity in Bert models does not offer a meaningful notion of sentence similarity~\cite{li2020sentence}, even performing worse than GLoVe embedding.

 \begin{table*}[t!]
    \centering% This is an environment - we probably don't want the extra spacing of center in addition to that added by table etc.
    \caption{AUC-DEL table for various methods Toxicity with no overlap and embedding not fixed. }
    \vspace{1mm}
    \label{tab:auc-fix-nofreeze}
    \small
    \begin{tabular}{p{0.08\textwidth} p{0.14\textwidth} p{0.07\textwidth}p{0.07\textwidth}p{0.10\textwidth}p{0.10\textwidth}p{0.10\textwidth}}
      \toprule% nicer rules courtesy of booktabs - but then we need to drop the verticals 
      Dataset & Metric & TR-last& TR-WE& TR-WE-topk & TR-WE-Syn& TR-WE-NoC \\
      \midrule
        Toxic  & AUC-DEL$+\downarrow$ & $\ \ \ 0.001$& $\ \ \ {0.002}$ & $\ \ \ 0.003$ & $\ \ \ 0.004$ & $\ \ \ 0.006$  \\
        Nooverlap & AUC-DEL$-\uparrow$ & $- 0.013$& $- 0.003$ & $- 0.007$ & $-0.008$ & $-0.004$ \\
      \bottomrule
      %\vspace{-5mm}
    \end{tabular}
    \vspace{-2mm}
  \end{table*}

\section{A Relaxation to Synonym Matching}
\vspace{-2mm}
While common tokens like ``start'' and ``end'' allow TracIn-WE to implicitly capture influence between sentences without word-overlap, the influence cannot be naturally decomposed over words in the two sentences. This hurts interpretability.
To remedy this, we propose a relaxation of TracIn-WE, called TracIn-WE-Syn, which allows for synonyms in two sentences to directly affect the influence score.
In what follows, we define synonyms to be words with similar embeddings.

We first rewrite word gradient similarity as
\begin{equation*}\label{eq:trackemb_we2}
    \begin{split}\small
    \vspace{-2mm}
        \text{WGS}_{x,x'}(w,w')  = \frac{\partial \ell(x, \Theta)}{\partial \Theta_{w}} ^T \frac{\partial \ell(x', \Theta)}{\partial \Theta_{w'}} \mathbbm{1}[w = w'].
    \vspace{-2mm}
    \end{split}
\end{equation*}
TracIn-WE can then be represented in the following form: 
{\small
\begin{equation*} \label{eq:trackemb_wo}
\begin{split}\small
    \text{TracIn-WE}(x,x') = - \sum_{w \in  x } \sum_{w' \in x'} \text{WGS}_{x,x'}(w,w').
\end{split}
\end{equation*}
}
which can be seen as the sum of word gradient similarities for matching words in the two sentences. 
It is then natural to consider the variant where exact match is relaxed to synonym match:
\begin{equation*}
    \begin{split}\small
    \vspace{-2mm}
        \text{WGS-syn}_{x,x'}(w,w')  = \frac{\partial \ell(x, \Theta)}{\partial \Theta_{w}} ^T \frac{\partial \ell(x', \Theta)}{\partial \Theta_{w'}} \mathbbm{1}[\text{Syn}(w,w') = 1].
    \vspace{-2mm}
    \end{split}
\end{equation*}
where Syn($w,w')=1$ if the cosine similarity of the embeddings of $w$ and $w'$ is above a threshold. We set the threshold to be $0.7$ in our experiments.
However, this direct relaxation has the caveat that a word $w$ in $x$ may be matched to several synonyms (including itself) in $x'$ simultaneously, which is not in the spirit of TracIn-WE where each word should only be matched to at most one word.
To resolve this, we seek an optimal 1:1 match between words between the two sentence that respects synonymy and maximizes influence.
We formulate this in terms of the Monge assignment problem~\citep{peyre2019computational} from optimal transport.
For scalability reasons, we operate on the top-$k$ relaxation of TracIn-WE (Section~\ref{sec:computational_tricks}).
%We thus leverage a common approach in optimal transport, and transform the matching problem into a so-called Monge assignment problem~\citep{peyre2019computational}, where each word in $x$ can only mapped to at most one word in $x'$. 
Let $\{w_1, w_2, ... w_k\}$ and $\{w'_1, w'_2, ... w'_k\}$ be the top-$k$ words contained in $x$ and $x'$ respectively.
Our goal is to find the optimal assignment
function $m \in \mathbb{M}: \{1, ..., k\} \rightarrow\{1, ..., k\}$, such that $m(i) \not = m(j)$ for $i \not = j$ where
%Given that $x$ and $x'$ words after pruning such that $x = \{w_1, w_2, ... w_k\}$, $x' = \{w'_1, w'_2, ... w'_k\}$, our goal is to find the optimal $1$-to-$1$ assignment function $m \in \mathbb{M}: \{1, ..., k\} \rightarrow\{1, ..., k\}$, such that $m(i) \not = m(j)$ for $i \not = j$ where 
\begin{equation} \label{eq:assignment}
\small
    m^* = \arg\min_{m \in \mathbb{M}}\sum_{i=1}^k -|\text{WGS-syn}_{x,x'}(w_i,w'_{m(i)})|.
\end{equation}
We define the matching cost between $w$ and $w'$ to be the negative absolute value of the word gradient similarity, as this allows us to match synonyms with strong positive as well as strong negative influence.
Optimal assignment can be calculated efficiently by existing solvers, for instance, linear\_sum\_assignment function in SKlearn~\citep{scikit-learn}.
The final total influence can be obtained by
\begin{equation*} \label{eq:trackemb-syn}
\begin{split}
\small
    \text{TracIn-WE-Syn}(x,x) =  -\sum_{w_i \in  x } \text{WGS-syn}_{x,x'}(w_i,w'_{m^*(i)}),
    \vspace{-6mm}
\end{split}
\end{equation*}

We report the result for this relaxation in the following table \ref{tab:auc-sup}, the result for TR-WE-Syn is close to the result of TracIn-WE, hinting that the additional synonym matching is not particular helpful for the deletion evaluation.

 \begin{table*}[t!]
    \centering% This is an environment - we probably don't want the extra spacing of center in addition to that added by table etc.
    \caption{AUC-DEL table for various methods (including TR-WE-Syn)in different datasets. Best number is bold.}
    \vspace{1mm}
    \label{tab:auc-sup}
    \small
    \adjustbox{max width=0.98\columnwidth}{
    \begin{tabular}{p{0.08\textwidth} p{0.14\textwidth} p{0.09\textwidth}p{0.07\textwidth}p{0.07\textwidth}p{0.08\textwidth}p{0.12\textwidth}p{0.13\textwidth}p{0.12\textwidth}}
      \toprule% nicer rules courtesy of booktabs - but then we need to drop the verticals 
      Dataset & Metric & Inf-Last & Rep& TR-last& TR-WE& TR-WE-topk &  TR-TFIDF &  TR-WE-Syn\\
      \midrule
      Toxic & AUC-DEL$+\downarrow$ & $-0.008$ & $-0.008$ & $-0.013$& $\mathbf{-0.100}$ & $-0.099$ & $-0.067$ &$\ \ \ 0.016$\\
      Bert & AUC-DEL$-\uparrow$ & $\ \ \ 0.014$ & $\ \ \ 0.021$ & $\ \ \ 0.023$& $\ \ \ 0.149$ & $\ \ \ \mathbf{0.151}$ & $\ \ \ 0.063$ & $\ \ \ 0.014$\\
      \midrule
      AGnews & AUC-DEL$+\downarrow$ & $-0.018$ & $-0.016$ & $-0.021$& $-0.166$ & $\mathbf{-0.174}$ & $-0.090$ & $-0.017$\\
      Bert & AUC-DEL$-\uparrow$ & $\ \ \ 0.033$ & $\ \ \ 0.028$ & $\ \ \ 0.028$& $\ \ \ {0.130}$ & $\ \ \ \mathbf{0.131}$ & $\ \ \  0.072$ & $\ \ \ 0.023$\\
      \midrule
       Dataset & Metric & Inf-Last & Rep& TR-last& TR-WE& TR-WE-topk & TR-WE-NoC & TR-common \\
       \midrule
        Toxic  & AUC-DEL$+\downarrow$ & $-0.009$ & $-0.008$ & $-0.006$& $\mathbf{-0.018}$ & $-0.016$ & $\ \ \ 0.003$ & $-0.008$ \\
        Nooverlap & AUC-DEL$-\uparrow$ & $ \ \ \ 0.008$ & $\ \ \ 0.007$ & $\ \ \ 0.010$& $\ \ \ \mathbf{0.026}$ & $\ \ \ \mathbf{0.026}$ & $\ \ \ 0.001$ & $\ \ \ 0.015$ \\
      \bottomrule
      %\vspace{-5mm}
    \end{tabular}}
    \vspace{-1mm}
  \end{table*}

\section{Qualitative Examples}

We show qualitative examples of the top-proponents and top-opponents for two random test points on dataset Toxicity (Tab.~\ref{tab:toxic_qual_last}, \ref{tab:toxic_qual_we}), AGnews (Tab.~\ref{tab:agnews_qual_last}, \ref{tab:agnews_qual_we}), and MNLI (Tab.~\ref{tab:mnli_qual_last}, \ref{tab:mnli_qual_we}).

 \begin{table*}[!t]
    \centering% This is an environment - we probably don't want the extra spacing of center in addition to that added by table etc.
    \caption{Proponents and opponents for TracIn-Last on Toxicity}
    \label{tab:toxic_qual_last}
    \small
    \begin{tabular}{p{0.12\textwidth} p{0.70\textwidth} p{0.1\textwidth}}
      \toprule
      & Sentence content & Label \\
      \toprule% nicer rules courtesy of booktabs - but then we need to drop the verticals 
      Test Sentence & I find Sandstein's dealing with the Mbz1 phenomenon very professional. He removed the soapbox image from that user's page and also banned you for not complying with your topic ban. It is you the one who is not assimilating the teaching of your topic ban. For example. You are topic banned because you don't have a professional approach to I-P topic and in general to any topic related to Jews and Judaism. The most resent example. When you reported that soapbox you qualified it as antisemitic. You at least should get informed of what that is. A neutral approach would be to have called it as soapbox canvasing and that's it. You should focus in your pictures which is the thing that you manage to do relatively well. Once you get into your holly war program of fighting all that in your imagination is an attack to Judaism you simply behave stupidly. It is those kinds of behaviors the ones that keep bringing hatred to us. That kind of attitude is, know it, racist, and if you are true to the struggles of the people of Abraham you above all should regret behaving as a racist. Once more, focus on your pictures and maybe even Sandstein will take a like on you.  & Non-Toxic  \\\midrule% Note that there is no & before the first column - & only comes between columns so if you define n columns, you can have at most n-1 & symbols in any row
      Proponent-1 & You mean my past BLOCK. The third block was because of your incompetence. Jesus doesn't like liars.  & Non-Toxic.\\% No need for 6 rows of space for each title!
      Proponent-2 & Pontiac Monrana 
Karrmann you full of shit ibelive all of the people who know that the montan will return after 2008
and we want ot knwo ehre do you get your info form and can you sohw it 
and guess what you dont know anythng about the Montana & Non-Toxic\\% No need for 6 rows of space for each title!
      \midrule
      Opponent-1 & I doubt this will get through your thick head (it's not an insult, it's an opinion based on your response) but the problem is not the issue itself. It's that people like you seem to enjoy (whether or not your side gets it right) to discuss, turn, twist and frankly abuse topics like this which are detrimental to the basic goals of Wikis in general and Wikipedia in particular. As John Stewart said to two hacks; You're hurting us.
& Toxic \\% No need for 6 rows of space for each
      Opponent-2 &Pss Rex, be sure to DOCUMENT all the things you've discovered
 on the John Kerry page etc.
 It's awesome that you and I INDEPENDENTLY observed (and can corrorborate) virtually
the exactsame  pattern by these liberals. Demonizing conservatives; lionizing liberals. It's repeated ad infinitum, ad nauseum.
The more proof we have, the easier it will be to persuade all but their fellow brain-dead truth haters to give a red cent to Wikipedia. And, until WHOLESALE changes are made from the top down, that's exactly what's about to happen. It's almost like this is the liberal's religion. Too bad they're gonna have to find a church other than Wikipedia to practice their faith, huh?
I've heard rumors that my actions are already sending users Hippocrite, Fred Bauder, WoohooKitty, Kizzle, FVW, Derex and especially the pimply faced 15 year old RedWolf
 to become so verklempt they don't know whether to schedule an appointement with their psychiatrist...or their gynecologist.
Big Daddy- PHASE II Dry up the funding  (on the road) & Toxic \\% No need for 6 rows of space for each title!
\midrule
      Test Sentence & Dear Mr.RHaworth 
We have relised that your wikipedia page has come under a number of violent attacks
We are happy with this as you are an extreme nob
Wikipedia advises that you imediantly kill yourself.
Thank you
- Wikipedia Staff  & Non-Toxic  \\\midrule% Note that there is no & before the first column - & only comes between columns so if you define n columns, you can have at most n-1 & symbols in any row
      Proponent-1 &Pontiac Monrana 
Karrmann you full of shit ibelive all of the people who know that the montan will return after 2008 and we want ot knwo ehre do you get your info form and can you sohw it 
and guess what you dont know anythng about the Montana  & Non-Toxic.\\% No need for 6 rows of space for each title!
      Proponent-2 & You mean my past BLOCK. The third block was because of your incompetence. Jesus doesn't like liars. & Non-Toxic\\% No need for 6 rows of space for each title!
      \midrule
      Opponent-1 & "
You are by far the most unhelpful, ungracious administrator I have ever had to deal with. You're incompetence is displayed in every encounter we have. Oh, and I'm quite familar with WP:NPA, which you resort to citing whenever you don't get your way. For other administrators who wish to be helpful, my last username was the Arabic version of Warraq. Warraq means ""scribe.""  
" & Toxic \\% No need for 6 rows of space for each
      Opponent-2 &"
Whoever you are, you tedious little twat, bombarding innocent users with these ""warnings"", realise that this IP address is shared by literally hunderds(and possibly thousands) of users, and the spammer(or spammers) represent less than 1 per cent of people posting/editing etc on this IP address. Unless you are just some dweeb who gets off on threatening people?"
 & Toxic \\% No need for 6 rows of space for each title!
\midrule
      \bottomrule
    \end{tabular}
    \vspace{-3mm}
  \end{table*}

 \begin{table*}[!t]
    \centering% This is an environment - we probably don't want the extra spacing of center in addition to that added by table etc.
    \caption{Proponents and opponents for TracIn-WE on toxicity}
    \label{tab:toxic_qual_we}
    \small
    \begin{tabular}{p{0.12\textwidth} p{0.60\textwidth} p{0.1\textwidth} p{0.1\textwidth}}
      \toprule
      & Sentence content & Label & Salient word \\
      \toprule% nicer rules courtesy of booktabs - but then we need to drop the verticals 
      Test Sentence & I find Sandstein's dealing with the Mbz1 phenomenon very professional. He removed the soapbox image from that user's page and also banned you for not complying with your topic ban. It is you the one who is not assimilating the teaching of your topic ban. For example. You are topic banned because you don't have a professional approach to I-P topic and in general to any topic related to Jews and Judaism. The most resent example. When you reported that soapbox you qualified it as antisemitic. You at least should get informed of what that is. A neutral approach would be to have called it as soapbox canvasing and that's it. You should focus in your pictures which is the thing that you manage to do relatively well. Once you get into your holly war program of fighting all that in your imagination is an attack to Judaism you simply behave stupidly. It is those kinds of behaviors the ones that keep bringing hatred to us. That kind of attitude is, know it, racist, and if you are true to the struggles of the people of Abraham you above all should regret behaving as a racist. Once more, focus on your pictures and maybe even Sandstein will take a like on you.
  & Non-Toxic  \\\midrule% Note that there is no & before the first column - & only comes between columns so if you define n columns, you can have at most n-1 & symbols in any row
      Proponent-1 &Hello, Mandarax, you put a banner in the talk of François Asselineau saying that the article may not respect the policy on biographies of living persons in the reason that is poorly sourced. All the sources (17) are coming from the French national press. & Non-Toxic & ass\\
      Proponent-2 & Hey Asspro 
I left a disparaging remark about you on the helpdesk. Enjoy. & Non-Toxic & ass\\
      \midrule
      Opponent-1 & "
Road Good aRticles are a pain in the ass. I've asked some one involved with the project what needs to be done with the article to take it to GA.  The history needs to be there, especially the early history which is missing.  It also needs more information on the road features, such as intersections, road width, road markings, etc. The few sources I've looked at haven't included this, and the roads people tend to use Google Maps as a source for the major intersections part.
" & Toxic & ass \\% No need for 6 rows of space for each
      Opponent-2 &
hi ho no more 
I will be systematically reviewing all of your edits with the description ""hi ho"" when you make an edit explain the edit you made so that the other editors can follow the progress or digress of the article instead of being a jack ass. & Toxic  & ass\\% No need for 6 rows of space for each title!
\midrule
      Test Sentence & Dear Mr.RHaworth 
We have relised that your wikipedia page has come under a number of violent attacks
We are happy with this as you are an extreme nob
Wikipedia advises that you imediantly kill yourself.
Thank you
- Wikipedia Staff  & Non-Toxic  \\\midrule% Note that there is no & before the first column - & only comes between columns so if you define n columns, you can have at most n-1 & symbols in any row
      Proponent-1 &. And to you you conservative being, you are a M Cosgrove defender. Admit it yourself
 & Non-Toxic & you\\% No need for 6 rows of space for each title!
      Proponent-2 & you want to know why it will kill manga well il tell you yours will be like a poison that will affect evreything in the world if you write your manga the stock market will crash the world will flood pepole will commit crimes arrmageddon will be rain down on us & Non-Toxic & kill\\% No need for 6 rows of space for each title!
      \midrule
      Opponent-1 & "
Please take time today to kill yourself.  We would all enjoy it.& Toxic & kill\\% No need for 6 rows of space for each
      Opponent-2 &"
CALTLAS IF YOU CONTINUE TO RVV IT, I WILL BLANK YOUR TALK PAGE, I WILL BLOCK YOU, I WILL BAN YOU, AND I WILL KILL YOUCALTLAS IF YOU CONTINUE TO RVV IT, I WILL BLANK YOUR TALK PAGE, I WILL BLOCK YOU, I WILL BAN YOU, AND I WILL KILL YOUCALTLAS IF YOU CONTINUE TO RVV IT, I WILL BLANK YOUR TALK PAGE, I WILL BLOCK YOU, I WILL BAN YOU, AND I WILL KILL YOUCALTLAS IF YOU CONTINUE TO RVV IT, I WILL BLANK YOUR TALK PAGE... (remove repetition)\\% No need for 6 rows of space for each title!
\midrule
      \bottomrule
    \end{tabular}
    \vspace{-3mm}
  \end{table*}

 \begin{table*}[!t]
    \centering% This is an environment - we probably don't want the extra spacing of center in addition to that added by table etc.
    \caption{Proponents and opponents for TracIn-Last on AGnews}
    \label{tab:agnews_qual_last}
    \small
    \begin{tabular}{p{0.12\textwidth} p{0.70\textwidth} p{0.1\textwidth}}
      \toprule
      & Sentence content & Label \\
      \toprule% nicer rules courtesy of booktabs - but then we need to drop the verticals 
     Test Sentence & Sheik Ahmed bin Hashr Al-Maktoum earned the first-ever Olympic medal for the United Arab Emirates when he took home the gold medal in men 39s double trap shooting on Tuesday in Athens. 
  & sports \\
  \midrule

      Proponent-1 & ARSENE WENGER is preparing for outright confrontation with the FA over his right to call Ruud van Nistelrooy a cheat. Arsenal boss Wenger was charged with improper conduct by Soho Square for his comments after   & Sport \\% No need for 6 rows of space for each title!
      Proponent-2 & AFP - Shaquille O'Neal paid various women hush money to keep quiet about sexual encounters, Kobe Bryant told law enforcement officers in Eagle, Colorado. & Sport\\% No need for 6 rows of space for each title!
      \midrule
      Opponent-1 & AFP - Jermain Defoe has urged Tottenham to snap up his old West Ham team-mate Joe Cole who is out of favour with Chelsea manager Jose Mourinho.
& World \\% No need for 6 rows of space for each
      Opponent-2 &AP - Democratic Party officials picked U.S. Rep. William Lipinski's son Tuesday to replace his father on the November ballot, a decision engineered by Lipinski after he announced his retirement and withdrew from the race four days earlier. & World \\% No need for 6 rows of space for each title!
\midrule
       Test Sentence & NEW YORK - Investors shrugged off rising crude futures Wednesday to capture well-priced shares, sending the Nasdaq composite index up 1.6 percent ahead of Google Inc.'s much-anticipated initial public offering of stock.    In afternoon trading, the Dow Jones industrial average gained 67.10, or 0.7 percent, to 10,039.93...
 & World  \\\midrule% Note that there is no & before the first column - & only comes between columns so if you define n columns, you can have at most n-1 & symbols in any row
      Proponent-1 & NEW YORK - Investors bid stocks higher Tuesday as oil prices declined and earnings results from a number of companies, including International Business Machines Corp. and Texas Instruments Inc., topped Wall Street's expectations...
 & World\\% No need for 6 rows of space for each title!
      Proponent-2 & NEW YORK - Investors bid stocks higher Tuesday as oil prices declined and earnings results from a number of companies, including International Business Machines Corp. and Texas Instruments Inc., topped Wall Street's expectations...
 & World\\% No need for 6 rows of space for each title!
      \midrule
      Opponent-1 & China protests against a US investigation that could lead a to trade war over China's cotton trouser trade. & Business \\% No need for 6 rows of space for each
      Opponent-2 &A new anti-corruption watchdog for Bangladesh has been welcomed by global anti-graft campaigners. & Business \\% No need for 6 rows of space for each title!
\midrule
      \bottomrule
    \end{tabular}
    \vspace{-3mm}
  \end{table*}

 \begin{table*}[!t]
    \centering% This is an environment - we probably don't want the extra spacing of center in addition to that added by table etc.
    \caption{Proponents and opponents for TracIn-WE on AGnews}
    \label{tab:agnews_qual_we}
    \small
    \begin{tabular}{p{0.12\textwidth} p{0.60\textwidth} p{0.1\textwidth} p{0.1\textwidth}}
      \toprule
      & Sentence content & Label & Salient word \\
      \toprule% nicer rules courtesy of booktabs - but then we need to drop the verticals 
      Test Sentence & Sheik Ahmed bin Hashr Al-Maktoum earned the first-ever Olympic medal for the United Arab Emirates when he took home the gold medal in men 39s double trap shooting on Tuesday in Athens. 
  & sports  \\\midrule% Note that there is no & before the first column - & only comes between columns so if you define n columns, you can have at most n-1 & symbols in any row
      Proponent-1 & ATHENS, Aug. 19 -- Worried about the potential for a terrorist catastrophe, Greece is spending about \$1.5 billion on security for the Olympic Games. The biggest threats so far? Foreign journalists and a Canadian guy dressed in a tutu.  & Sports & olympic\\
      Proponent-2 & ATHENS (Reuters) - A Canadian man advertising an online gaming site, who broke security and jumped into the Olympic diving pool, has been given a five-month prison term for trespassing and disturbing public order, court officials say.  & Sports & olympic\\
      \midrule
      Opponent-1 & "
Britain's Kelly Holmes storms to a sensational Olympic 800m gold in Athens.
" & World & olympic \\% No need for 6 rows of space for each
      Opponent-2 &
AFP - Britain were neck and neck with Olympic minnows Slovakia and Zimbabwe and desperately hoping for an elusive gold medal later in the week. & World  & olympic\\% No need for 6 rows of space for each title!
\midrule
      Test Sentence & NEW YORK - Investors shrugged off rising crude futures Wednesday to capture well-priced shares, sending the Nasdaq composite index up 1.6 percent ahead of Google Inc.'s much-anticipated initial public offering of stock.    In afternoon trading, the Dow Jones industrial average gained 67.10, or 0.7 percent, to 10,039.93...
 & World  \\\midrule% Note that there is no & before the first column - & only comes between columns so if you define n columns, you can have at most n-1 & symbols in any row
      Proponent-1 &. NEW YORK - Stocks are seen moving lower at the open Wednesday as investors come to grips with the Federal Reserve hiking its key rates by a quarter point to 1.75 percent.    Dow Jones futures fell 14 points recently, while Nasdaq futures were down 2.50 points and S P futures dropped 1.80 points...
& World & futures\\% No need for 6 rows of space for each title!
      Proponent-2 &NEW YORK - Stocks were little changed early Wednesday as investors awaited testimony from Federal Reserve Chairman Alan Greenspan before a House budget panel.    In morning trading, the Dow Jones industrial average was down 0.08 at 10,342.71...
 & World& investors\\% No need for 6 rows of space for each title!
      \midrule
      Opponent-1 & Google Saves Kidnapped Journalist in Iraq Google can claim another life saved after a kidnapped Australian journalist was freed by his captors in Iraq earlier today. Freelance journalist John Martinkus was abducted by gunmen on Saturday outside a hotel near the Australian embassy. Apparently Martinkus was able to convince his captors ...& Sci/Tech & google\\% No need for 6 rows of space for each
      Opponent-2 & With a 9:15 p.m. curfew imposed because of Hurricane Jeanne, Tampa Bay beat Toronto with 39 minutes to spare. Hoping to beat the storm, the Blue Jays were scheduled to leave Florida on a charter flight immediately after the loss. Today's series finale was canceled because of the hurricane, which was expected to hit Florida's east coast late yesterday or ...
 & sport & `.' \\% No need for 6 rows of space for each title!
\midrule
      \bottomrule
    \end{tabular}
    \vspace{-3mm}
  \end{table*}

 \begin{table*}[!t]
    \centering% This is an environment - we probably don't want the extra spacing of center in addition to that added by table etc.
    \caption{Proponents and opponents for TracIn-Last on MNLI}
    \label{tab:mnli_qual_last}
    \small
    \begin{tabular}{p{0.12\textwidth} p{0.70\textwidth} p{0.1\textwidth}}
      \toprule
      & Sentence content & Label \\
      \toprule% nicer rules courtesy of booktabs - but then we need to drop the verticals 
     Test Sentence & Premise: To some critics, the mystery isn't, as Harris suggests, how women throughout history have exploited their sexual power over men, but how pimps like him have come away with the profit.
     
     Hypothesis:
     Harris suggests that it's a mystery how women have exploited men with their sexual power.
  & Entailment \\
  \midrule

      Proponent-1 & Premise:Also in Back Lane are the headquarters of An Taisce, an organization dedicated to the preservation of historic buildings and gardens. Hypothesis: The headquarters of An Taisce are located in Black Lane.& Entailment\\% No need for 6 rows of space for each
      Proponent-2 & Premise: yeah you know because they they told us in school that you know crime has to be an intent you know has to be not just the act but you have to intend to do it because there could be accidental kind of things you know. Hypothesis:I was told in school that if you do something bad by accident it is not a crime.
& Entailment \\
      \midrule
      Opponent-1 & Premise: I still can't quite believe that. Hypothesis:I don't believe that at all.
& Contradiction \\% No need for 6 rows of space for each
      Opponent-2 & Premise: The problem isn't so much that men are designed by natural selection to fight as what they're designed to fight women . Hypothesis: Women were designed by natural selection to fight men.
& Contradiction \\% No need for 6 rows of space for each
\midrule 

Test Sentence & Premise:Mykonos has had a head start as far as diving is concerned because it was never banned here (after all, there are no ancient sites to protect)
     
     Hypothesis:
    Diving was banned in places other than Mykonos.
  & Entailment \\
  \midrule

      Proponent-1 & Premise:yeah i could use a discount  i have to wait for the things to go on sale. Hypothesis: I wait for sales now, and it's very convenient.& Entailment \\% No need for 6 rows of space for each
      Proponent-2 & Premise: you know and then we have that you know if you can't stay if something comes up and you can't stay within it then we have uh you know a budget for you know like we call our slush fund or something and something unexpected unexpected comes up then you're not. Hypothesis: Having a slush fund helps to pay for things that are not in the budget in case of emergencies. 
& Entailment \\
      \midrule
      Opponent-1 & Premise: Farrow is humorless and steeped in a bottomless melancholy. Hypothesis: Farrow is depressed and acting very sad.& Neutral \\% No need for 6 rows of space for each
      Opponent-2 & Premise: Julius leaned forward, and in doing so the light from the open door lit up his face. Hypothesis: Julius moved so that the light could illuminate his face.
& Neutral \\% No need for 6 rows of space for each
\midrule
      \bottomrule
    \end{tabular}
    \vspace{-3mm}
  \end{table*}

 \begin{table*}[!t]
    \centering% This is an environment - we probably don't want the extra spacing of center in addition to that added by table etc.
    \caption{Proponents and opponents for TracIn-WE-topk on MNLI}
    \label{tab:mnli_qual_we}
    \small
    \begin{tabular}{p{0.12\textwidth} p{0.60\textwidth} p{0.1\textwidth} p{0.1\textwidth}}
      \toprule
      & Sentence content & Label & Salient Word \\
      \toprule% nicer rules courtesy of booktabs - but then we need to drop the verticals 
     Test Sentence & Premise: To some critics, the mystery isn't, as Harris suggests, how women throughout history have exploited their sexual power over men, but how pimps like him have come away with the profit.
     
     Hypothesis:
     Harris suggests that it's a mystery how women have exploited men with their sexual power.
  & Entailment \\
  \midrule

      Proponent-1 & Premise: but get up during every commercial and things like that and you'd be surprised at how much just that little bit adds up you know just gives you a little more activity so. Hypothesis: You won't get any significant exercise by moving around during commercial breaks.& Contradiction & `'t'\\% No need for 6 rows of space for each
      Proponent-2 & Premise: From Chapter 4, a 500 MWe facility will need about 175 tons of steel to install an ACI system, or about 0.35 tons per MWe. Hypothesis: A 500 MWe needs steel to install an ACI system
& Entailment & [end] \\
      \midrule
      Opponent-1 & Premise: Also exhibited are examples of Linear B type, which was deciphered in 1952 and is of Mycenaean origin showing that by the time the tablet was written the Minoans had lost control of the major cities. Hypothesis: Although Linear B has been deciphered, Linear A is still a mystery.& Contradiction & Mystery \\% No need for 6 rows of space for each
      Opponent-2 & Premise: The problem isn't so much that men are designed by natural selection to fight as what they're designed to fight women . Hypothesis: Women were designed by natural selection to fight men. 
& Entailment & women \\% No need for 6 rows of space for each
\midrule 

Test Sentence & Premise:Mykonos has had a head start as far as diving is concerned because it was never banned here (after all, there are no ancient sites to protect)
     
     Hypothesis:
    Diving was banned in places other than Mykonos.
  & Entailment \\
  \midrule

      Proponent-1 & Premise:and they have a job in jail and they work that they should i and this may sound cruel but i do not think that they should be allowed cigarettes i mean they're in jail for crying out loud what do they need cigarettes for. Hypothesis: I think cigarettes should be banned in prison.& Entailment & banned\\% No need for 6 rows of space for each
      Proponent-2 & Premise: If I fill in my name and cash it, I pay tax. Hypothesis:I'll have to pay taxes when I cash the check.
& Neutral & [end] \\
      \midrule
      Opponent-1 & Premise: Already, [interleague play] has restored one of baseball's grandest  the passion for arguing about the game, observed the Chicago Tribune . Things could be  The Los Angeles Times reports that, thanks to the popularization of baseball in Poland, bats have emerged as a weapon of choice for hooligans, thugs, [and] extortionists. Hypothesis: Baseball bats have been banned in Poland.& Neutral & banned\\% No need for 6 rows of space for each
      Opponent-2 & Premise: Because of the possible toxicity of thiosulfate to test organisms, a control lacking thiosulfate should be included in toxicity tests utilizing thiosulfate-dechlorinated water. Hypothesis: Because of the possible toxicity of thiosulfate to test organisms, it should be banned.
& Neutral & banned \\% No need for 6 rows of space for each
\midrule
      \bottomrule
    \end{tabular}
    \vspace{-3mm}
  \end{table*} 
  
\begin{figure*}
    \centering
    \begin{minipage}{0.5\textwidth}
        \centering
        \vspace{-1.6mm}
        \includegraphics[width=0.8\textwidth]{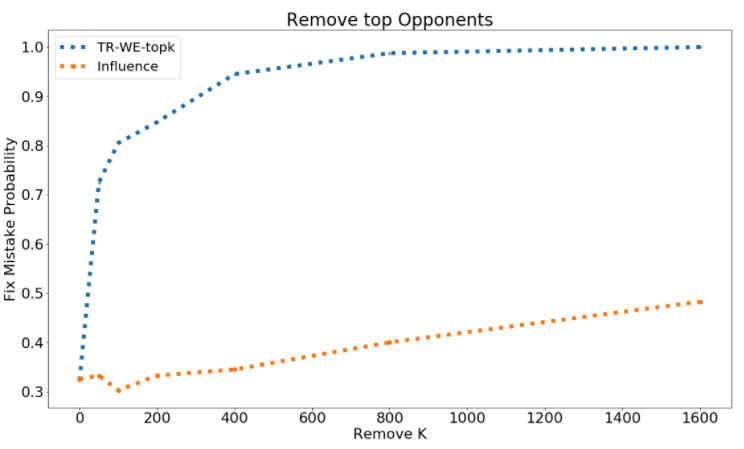} % first figure itself
        \vspace{-1mm}
    %\caption{Deletion curve for removing top opponents (larger is better).}
    \end{minipage}\hfill
    \begin{minipage}{0.5\textwidth}
        \centering
        \vspace{0.15mm}
        \includegraphics[width=0.8\textwidth]{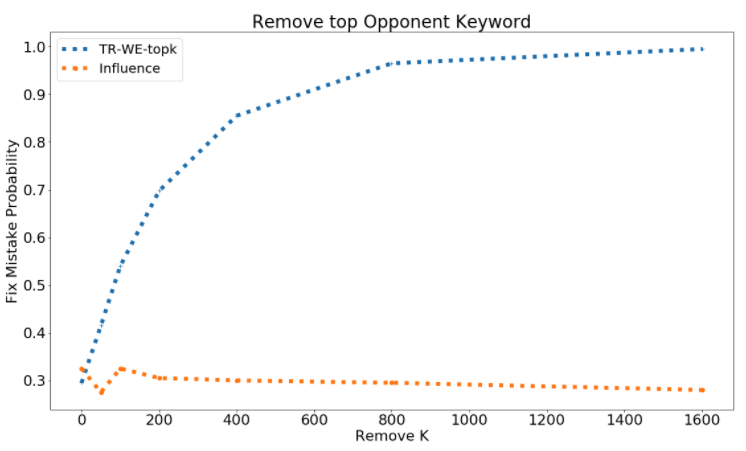} % second figure itself
         \vspace{-0.67mm}
        %\caption{Deletion curve for removing top proponents (smaller is better).}
    \end{minipage}
    \vspace{-1.0mm}
    \caption{ Probability to fix a mistake on Toxicity dataset by removing opponents and the removing one key word in opponents.}
    \label{fig:prob_toxic_fix}
    \vspace{-3mm}
\end{figure*}

\begin{figure*}
    \centering
    \begin{minipage}{0.5\textwidth}
        \centering
        \vspace{-1.6mm}
        \includegraphics[width=0.8\textwidth]{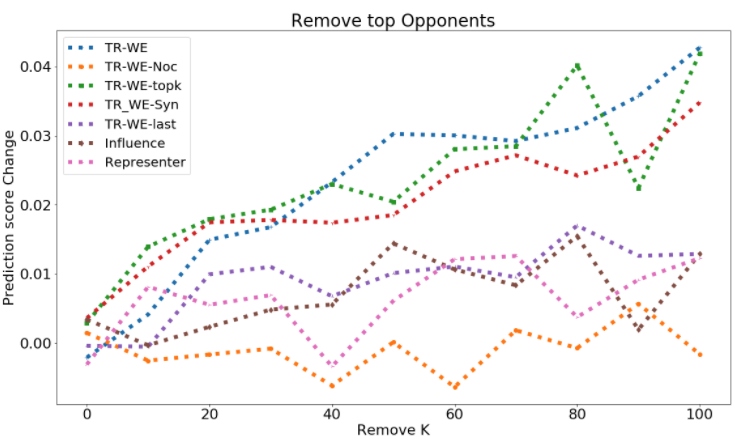} % first figure itself
        \vspace{-1mm}
    %\caption{Deletion curve for removing top opponents (larger is better).}
    \end{minipage}\hfill
    \begin{minipage}{0.5\textwidth}
        \centering
        \vspace{0.15mm}
        \includegraphics[width=0.8\textwidth]{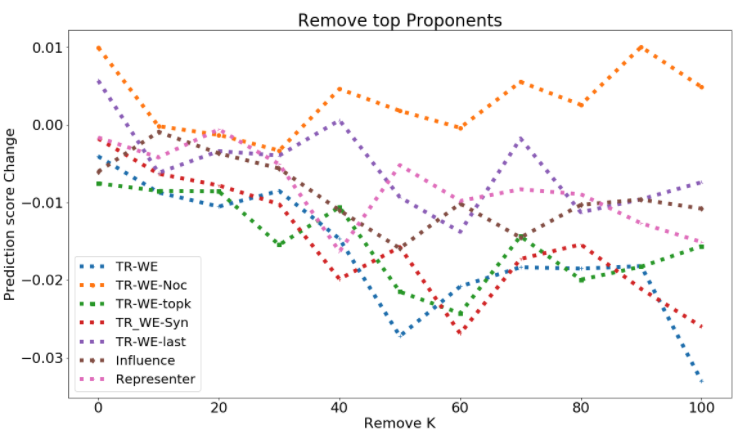} % second figure itself
         \vspace{-0.67mm}
        %\caption{Deletion curve for removing top proponents (smaller is better).}
    \end{minipage}
    \vspace{-1.0mm}
    \caption{Deletion Curve on Toxicity dataset for removing opponents (larger better) and the removing proponents (smaller better).}
    \label{fig:del_toxic_noo_fix}
    \vspace{-3mm}
\end{figure*}

\begin{figure*}
    \centering
    \begin{minipage}{0.5\textwidth}
        \centering
        \vspace{-1.6mm}
        \includegraphics[width=0.8\textwidth]{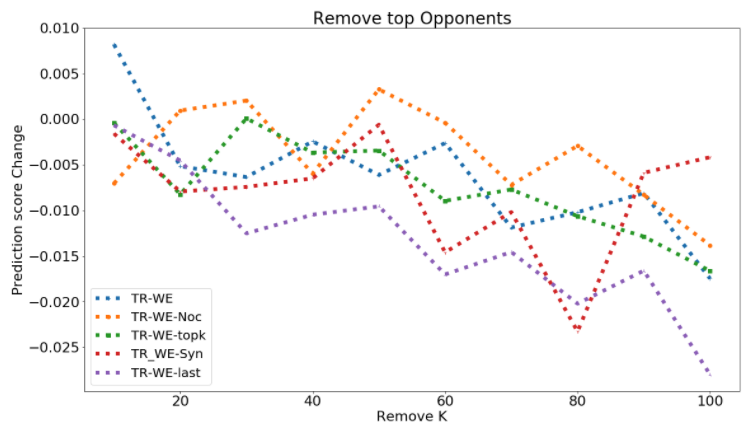} % first figure itself
        \vspace{-1mm}
    %\caption{Deletion curve for removing top opponents (larger is better).}
    \end{minipage}\hfill
    \begin{minipage}{0.5\textwidth}
        \centering
        \vspace{0.15mm}
        \includegraphics[width=0.8\textwidth]{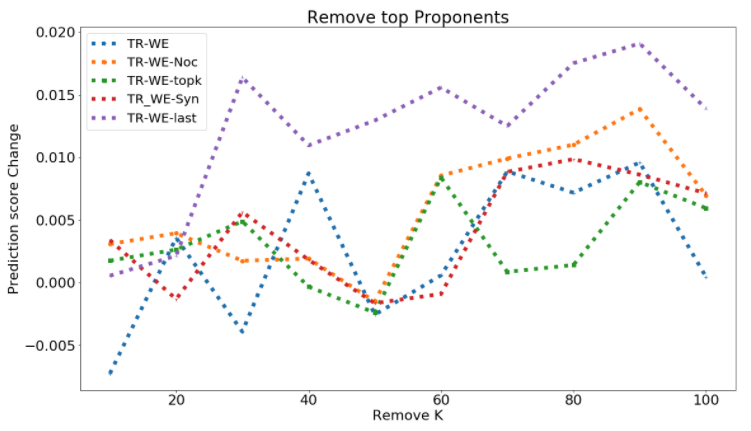} % second figure itself
         \vspace{-0.67mm}
        %\caption{Deletion curve for removing top proponents (smaller is better).}
    \end{minipage}
    \vspace{-1.0mm}
    \caption{Deletion Curve or Toxicity dataset for removing opponents (larger better) and the removing proponents (smaller better).}
    \label{fig:del_toxic_noo_nofix}
    \vspace{-3mm}
\end{figure*}

\section{No word overlap Experiment -- More Details}

\paragraph{Why Fix Word Embedding:}
We first start by the conclusion of our observation: many influence methods cannot find training examples that influences a test point without word overlap in the case where word embedding is not fixed. To support this observation, we show the deletion curve on no word overlap experiment (when word embedding is not fixed during training) in Fig. \ref{fig:del_toxic_noo_nofix} and the AUC-DEL score in Tab.~\ref{tab:auc-fix-nofreeze}. We can see after that removing proponents the Deletion score is actually slightly positive for all methods, and that removing opponents the Deletion score is actually slightly negative for all methods. This shows that no influence methods is able to find training examples that influence the test point without having word overlaps. 

We thus suspect that influence may flow through examples pairs without word overlaps when the embedding is fixed. The intuition is that if you have two words A and A', that have the same initial word embedding. When embedding is not fixed, the embedding of A' and A may grow apart during training. However, if the embedding is fixed, the input of A and A' will always be the same regardless of whether if the training is applied on A and A'. Based on this intuition, we fix the word embedding during the model training for the no word overlap experiment. We now show the deletion curve for our experiment on no word overlap (when word embedding is fixed during training) in Fig. \ref{fig:del_toxic_noo_fix} (which is omitted from main text due to space constraint). We observe that although the signal is weak, most methods other than TR-WE-Noc is consistently positive when opponents are removed, and consistently negative when proponents are removed. As our result of AUC-DEL suggests, TR-WE variants perform the best in this case.

\section{Other Experiment Details}

For Toxicity and AGnews, we use the small-Bert model\footnote{https://huggingface.co/google/bert\_uncased\_L-2\_H-128\_A-2} as our base model and fine-tune on our validation set. For Toxicity Roberta, we use the standard Roberta-Base \footnote{https://huggingface.co/roberta-base}.

For MNLI, we use normal Bert models\footnote{https://huggingface.co/bert-base-uncased} and fine-tune on the validation set. For checkpoint selection, we follow suggestions in~\citet{pruthi2020estimating} and choose $3-5$ checkpoints where the loss has not saturated yet. We follow standard fine-tuning procedures using SGD optimizers with momentum $0.9$, and we fine-tune for $10$ epochs on AGnews with $2e^{-2}$ learning rate and fine-tune for $20$ epochs on Toxicity with $2e^{-4}$ learning rate. The retraining parameters is fixed during the calculation of deletion curve. We split the training and validation set randomly ($50000$ training and $20000$ validation and $20000$ testing) and fix the random seed.

For MNLI, we calculate deletion curve for k $ \in [20,40,60,80,100,200,400,600,800,1000,5000]$, and we can see from Fig. \ref{fig:del_pro} that removing $60$ examples based on TracIn-WE-Syn affects the test point more than removing $5000$ examples based on TracIn-Last. %The fine-tuning of MNLI follows standard framework in Tensorflow model garden~\citep{tensorflowmodelgarden2020}.

We also clarify that in the context of our work, we refer to the tokens and words interchangeably for presentation simplicity. In our work, we use the tokenizer that is used along with Bert or Roberta, which contains mostly words but also some word piece. When using a character-based tokenizer, the usage of ``word'' would then become characters.

\section{Targeted Fixing of Misclassifications} \label{sec:application}
\vspace{-1mm}
We now discuss an application of our influence method in fixing specific misclassifications made by the model.
%In this section, we target a more direct fixing application in toxicity dataset, where our goal is to fix a specific mistake made by a model by modifying the training set. 
We propose two means of fixing (a) remove top-k opponents (b) replace the most negatively influential word in each of the top-k opponents by [PAD]. The most influential word may be identified using the word-level decomposition of TracIn-WE; see Section~\ref{sec:word_decomposition}.
We consider a BERT model for the toxicity comment classification task~\cite{toxic2018kaggle}, and randomly chose $40$ misclassifications from the test set with prediction probability in $[0.3,0.7]$. 
For each misclassification, we apply the two approaches mentioned above for various values of $k$.
For each $k$, we report the average percentage of the mistake being fixed in $10$ rounds of retraining.
%For each mistake, we either remove top-$k$ opponents or the most negatively influential key word in the top-$k$ opponents, and perform $10-$runs retraining, and report the average percentage of the mistake being fixed. 

We compare TracIn-WE-Topk with Influence-Last.
To identify the most influential word using Influence-Last, we consider its gradient w.r.t. to each word, which is suggested in a similar use case by \citet{pezeshkpour2021combining}.
 %Influence-Last and its training feature gradient (which is suggested in a similar use case by \citet{pezeshkpour2021combining}). 
For fix method (a), removing $50$ opponents by TracIn-WE-Topk can fix a mistake $73 \%$ of the time, while removing $50$ opponents by Influence-last can only fix it $33 \%$ of the time.
With both methods, the average accuracy of the model after removing $50$ examples only drops by $0.01\%$.
For fix method (b), removing the most negatively influential word for the top-$200$ opponents by TracIn-WE-Topk can fix a mistake $70 \%$ of the time, while the same for Influence-Last can only fix a mistake $30 \%$ of the time.
With both methods, the average accuracy of the model after removing the most negatively influential word in the top-$200$ opponents drops by less than $0.02\%$.

We show the full fixing curve in the fixing application in Fig~\ref{fig:prob_toxic_fix}, where x-axis is the number of training sentence we remove (either full remove or only remove one top key word). We show that TrackIn-WE-topk significantly outperforms Influence-last in the targeted fixing application across different number of removal $k$. When remove num $k=0$, we see that the fix probability is $0.3$, meaning that after direct retrain without removal, the mistake can actually be correctly classified by the model $30 \%$ of the time.

\end{document}